\documentclass[
paper=letter,%
numbers=noendperiod,%
captions=nooneline,%
fontsize=11pt,%
DIV=12%
]{scrartcl}

\newif\ifunblinded%
\unblindedtrue

\usepackage[T1]{fontenc}%
\usepackage{lmodern}%
\usepackage[american]{babel}%
\usepackage{microtype}%

\usepackage[
hyperref,%
table%
]{xcolor}%

\usepackage{scrlayer-scrpage}%

\usepackage{calc}%
\usepackage{rotating}%
\usepackage{amsmath}%
\usepackage{mathtools}%
\usepackage{amssymb}%
\usepackage{amsthm}%
\usepackage{thmtools}%
\usepackage{etoolbox}%
\usepackage{xparse}%
\usepackage{bm}%
\usepackage{bbm}%
\usepackage{enumitem}%
\usepackage[calc]{datetime2}%
\usepackage{graphicx}%
\usepackage{grffile}%
\usepackage{tikz}%
\usepackage{wrapfig}%
\usepackage{tabularx}%
\usepackage{siunitx}%
\usepackage{booktabs}%
\usepackage{multirow}%
\usepackage{vruler}%
\usepackage{fancyvrb}%
\usepackage{textcomp}%
\usepackage{listings}%
\usepackage{csquotes}%
\usepackage{subcaption}%
\usepackage[
style=authoryear,%
dashed=false,%
hyperref=true,%
useprefix=true,%
maxnames=2,%
uniquelist=minyear,%
maxbibnames=6,%
uniquename=false,%
seconds=true,%
date=iso,%
urldate=iso%
]{biblatex}%
\usepackage[
hypertexnames=false,%
setpagesize=false,%
pdfborder={0 0 0},%
pdfstartview=Fit,%
bookmarksopen=true,%
bookmarksnumbered=true%
]{hyperref}%
\usepackage[hyphens]{xurl}
\hypersetup{breaklinks=true}
\setkomafont{pageheadfoot}{\normalfont\normalcolor\sffamily}%
\setkomafont{pagenumber}{\normalfont\normalcolor\sffamily}%
\automark{section}%
\setkomafont{captionlabel}{\normalfont\normalcolor\sffamily\bfseries}%

\newcommand*{\mysquare}{\rule[0.18em]{0.36em}{0.36em}}
\newcommand*{\mytriangle}{\raisebox{0.12em}{\resizebox{0.48em}{0.48em}{$\blacktriangleright$}}}
\newcommand*{\mybar}{\rule[0.32em]{0.62em}{0.08em}}
\newcommand*{\mydot}{\raisebox{0.14em}{\resizebox{0.44em}{!}{$\bullet$}}}
\setlist{%
  align=left,%
  labelindent=0mm, %
  leftmargin=!,%
  itemindent=0mm, %
  listparindent=\parindent,%
  parsep=0mm,%
  topsep=1mm,%
  itemsep=1mm%
}
\setlist[itemize,1]{label={\mysquare\ }, labelwidth=\widthof{\mysquare\ }}%
\setlist[itemize,2]{label={\mytriangle\ }, labelwidth=\widthof{\mytriangle\ }}%
\setlist[itemize,3]{label={\mybar\ }, labelwidth=\widthof{\mybar\ }}%
\setlist[itemize,4]{label={\mydot\ }, labelwidth=\widthof{\mydot\ }}%
\setlist[enumerate,1]{label=\arabic*), labelwidth=\widthof{9)}}%
\setlist[enumerate,2]{label=\arabic{enumi}.\arabic*), labelwidth=\widthof{9.9)}}%
\setlist[enumerate,3]{label=\arabic{enumi}.\arabic{enumii}.\arabic*), labelwidth=\widthof{9.9.9)}}%
\setlist[enumerate,4]{label=\arabic{enumi}.\arabic{enumii}.\arabic{enumiii}.\arabic*), labelwidth=\widthof{9.9.9.9)}}%

\allowdisplaybreaks%
\newcommand*{\abstractnoindent}{}%
\let\abstractnoindent\abstract
\renewcommand*{\abstract}{\let\quotation\quote\let\endquotation\endquote
  \abstractnoindent}
\deffootnote[1em]{1em}{1em}{\textsuperscript{\thefootnotemark}}%
\definecolor{blue}{RGB}{58, 95, 205}%
\definecolor{red}{RGB}{205, 41, 144}%
\definecolor{orange}{RGB}{238, 118, 0}%
\definecolor{chocolate}{RGB}{205, 102, 29}%

\lstdefinestyle{input}{
  backgroundcolor=\color{black!12},%
  commentstyle=\itshape\color{black!50},%
  keywordstyle=\bfseries\color{black},%
  stringstyle=\color{black}%
}

\lstdefinestyle{output}{
  backgroundcolor=\color{black!6}%
}

\lstdefinestyle{codestyle}{
  language={},%
  keywords={},%
  otherkeywords={}%
}

\lstnewenvironment{codeinput}[1][]{%
  \lstset{style=input, style=codestyle}
  #1%
}{\vspace{-0.25\baselineskip}}%

\lstnewenvironment{codeoutput}[1][]{%
  \lstset{style=output, style=codestyle}
  #1%
}{\vspace{-0.25\baselineskip}}%

\expandafter\let\csname Sinput\endcsname\relax
\expandafter\let\csname endSinput\endcsname\relax
\expandafter\let\csname Soutput\endcsname\relax
\expandafter\let\csname endSoutput\endcsname\relax

\lstdefinestyle{Rstyle}{
  language=R,%
  keywords={},%
  otherkeywords={}%
}

\lstnewenvironment{Sinput}[1][]{%
  \lstset{style=input, style=Rstyle}
  #1%
}{\vspace{-0.25\baselineskip}}%

\lstnewenvironment{Soutput}[1][]{%
  \lstset{style=output, style=Rstyle}
  #1%
}{\vspace{-0.25\baselineskip}}%

\lstdefinestyle{Cstyle}{
  language=C,%
  keywords={},%
  otherkeywords={}%
}

\lstnewenvironment{Cinput}[1][]{%
  \lstset{style=input, style=Cstyle}
  #1%
}{\vspace{-0.25\baselineskip}}%

\lstnewenvironment{Coutput}[1][]{%
  \lstset{style=output, style=Cstyle}
  #1%
}{\vspace{-0.25\baselineskip}}%

\lstdefinestyle{Bashstyle}{
  language=bash,%
  keywords={},%
  otherkeywords={}%
}

\lstnewenvironment{Bashinput}[1][]{%
  \lstset{style=input, style=Bashstyle}
  #1%
}{\vspace{-0.25\baselineskip}}%

\lstnewenvironment{Bashoutput}[1][]{%
  \lstset{style=output, style=Bashstyle}
  #1%
}{\vspace{-0.25\baselineskip}}%

\lstdefinestyle{LaTeXstyle}{
  language=[LaTeX]TeX,%
  texcs={},%
  keywords={},%
  otherkeywords={}%
}

\lstnewenvironment{LaTeXinput}[1][]{%
  \lstset{style=input, style=LaTeXstyle}
  #1%
}{\vspace{-0.25\baselineskip}}%

\lstnewenvironment{LaTeXoutput}[1][]{%
  \lstset{style=output, style=LaTeXstyle}
  #1%
}{\vspace{-0.25\baselineskip}}%

\DeclareNameAlias{sortname}{family-given}%
\DefineBibliographyExtras{american}{\DeclareQuotePunctuation{}}%
\renewbibmacro*{volume+number+eid}{%
  \setunit*{\addcomma\space}%
  \printfield{volume}%
  \printfield{number}}
\DeclareFieldFormat*{number}{(#1)}
\DeclareFieldFormat*{title}{#1}%
\DeclareFieldFormat{doi}{%
  \ifhyperref
    {\href{http://dx.doi.org/#1}{\nolinkurl{doi:#1}}}%
    {\nolinkurl{doi:#1}}}%
\renewbibmacro*{in:}{}%
\DeclareFieldFormat{isbn}{ISBN #1}%
\DeclareFieldFormat{pages}{#1}%
\DeclareFieldFormat{url}{\url{#1}}%
\DeclareFieldFormat{urldate}{\mkbibparens{#1}}%
\renewcommand*{\cite}[2][]{\textcite[#1]{#2}}%

\newif\ifstarttheorem
\declaretheoremstyle[%
  spaceabove=0.5em,
  spacebelow=0.5em,
  headfont=\sffamily\bfseries\global\starttheoremtrue,
  notefont=\sffamily\bfseries,
  notebraces={(}{)},
  headpunct={},
  bodyfont=\normalfont,
  postheadspace=\newline%
]{myMainStyle}
\declaretheorem[style=myMainStyle, numberwithin=section]{definition}%

\declaretheorem[style=myMainStyle, sibling=definition]{algorithm}

\makeatletter
\preto\itemize{%
  \if@inlabel
    \ifstarttheorem
      \mbox{}\par\nobreak\vskip\glueexpr-\parskip-\baselineskip+0.25em\relax\hrule\@height\z@
    \fi%
  \fi%
  \global\starttheoremfalse%
 \def\tempa{proof}%
 \ifx\tempa\mycurrenvir
    \ifstarttheorem
      \mbox{}\par\nobreak\vskip\glueexpr-\parskip-\baselineskip+0.25em\relax\hrule\@height\z@
    \fi%
 \fi%
 \global\starttheoremfalse%
}
\preto\enditemize{\global\starttheoremfalse}
\makeatother

\makeatletter
\preto\enumerate{%
  \if@inlabel
    \ifstarttheorem
      \mbox{}\par\nobreak\vskip\glueexpr-\parskip-\baselineskip+0.25em\relax\hrule\@height\z@
    \fi%
  \fi%
  \global\starttheoremfalse%
 \def\tempa{proof}%
 \ifx\tempa\mycurrenvir
    \ifstarttheorem
      \mbox{}\par\nobreak\vskip\glueexpr-\parskip-\baselineskip+0.25em\relax\hrule\@height\z@
    \fi%
 \fi%
 \global\starttheoremfalse%
}
\preto\endenumerate{\global\starttheoremfalse}
\makeatother

\makeatletter
\NewDocumentCommand{\tmb}{O{0.1mm} O{0.1mm} O{0.88} m m m}{%
  \mathrel{%
    \vbox{\offinterlineskip\m@th
      \ialign{%
        \hfil##\hfil\cr
        $\scriptscriptstyle\text{\scalebox{#3}{#4}}\mathstrut$\cr%
        \noalign{\vspace{#1}}%
        \vtop{%
          \ialign{%
            \hfil##\hfil\cr
            $#5$\cr\noalign{\vspace{#2}}%
            $\scriptscriptstyle\text{\scalebox{#3}{#6}}\mathstrut$\cr%
          }%
        }\cr
      }%
    }%
  }%
}
\makeatother

\makeatletter
\NewDocumentCommand{\tmbc}{O{0.1mm} O{0.1mm} O{0.88} m m m}{%
  \mathrel{%
    \vbox{\offinterlineskip\m@th
      \ialign{%
        \hfil##\hfil\cr
        $\scriptscriptstyle\mathclap{\text{\scalebox{#3}{#4}}}\mathstrut$\cr%
        \noalign{\vspace{#1}}%
        \vtop{%
          \ialign{%
            \hfil##\hfil\cr
            $#5$\cr\noalign{\vspace{#2}}%
            $\scriptscriptstyle\mathclap{\text{\scalebox{#3}{#6}}}\mathstrut$\cr%
          }%
        }\cr
      }%
    }%
  }%
}
\makeatother

\usetikzlibrary{shapes.geometric}

\newcommand*{\T}{^{\top}}

\newcommand*{\IN}{\mathbb{N}}

\newcommand*{\IR}{\mathbb{R}}

\newcommand*{\LN}{\operatorname{LN}}

\newcommand*{\U}{\operatorname{U}}

\newcommand*{\N}{\operatorname{N}}

\newcommand*{\ARMA}{\operatorname{ARMA}}
\newcommand*{\GARCH}{\operatorname{GARCH}}

\newcommand*{\I}{\mathbbm{1}}
\newcommand*{\rd}{\mathrm{d}}

\renewcommand*{\P}{\mathbb{P}}
\newcommand*{\E}{\mathbb{E}}

\newcommand*{\bias}{\operatorname{bias}}
\newcommand*{\VRF}{\operatorname{VRF}}

\newcommand*{\ES}{\operatorname{ES}}

\newcommand*{\AC}{\operatorname{AC}}

\newcommand*{\MMD}{\operatorname{MMD}}
\newcommand*{\ACvM}{\operatorname{ACvM}}

\newcommand*{\R}{\textsf{R}}
\newcommand*{\eps}{\varepsilon}

\newcommand*{\ntrn}{n_{\text{trn}}}
\newcommand*{\ntst}{n_{\text{tst}}}
\newcommand*{\nbat}{n_{\text{bat}}}%
\newcommand*{\nbts}{n_{\text{bts}}}%
\newcommand*{\nepo}{n_{\text{epo}}}
\newcommand*{\nphs}{n_{\text{phs}}}
\newcommand*{\ngen}{n_{\text{gen}}}
\newcommand*{\krn}{\text{krn}}
\newcommand*{\nkrn}{n_{\krn}}
\newcommand*{\nrep}{n_{\text{rep}}}
\newcommand*{\ndat}{n_{\text{dat}}}
\newcommand*{\dpri}{d_{\text{pri}}}
\newcommand*{\hval}{\bm{h}_{\text{val}}}
\newcommand*{\STOP}{\operatorname{STOP}}

\newcommand*{\AGMMN}{\operatorname{AGMMN}}
\newcommand*{\mucopMC}{\hat{\mu}_{\ngen}^{C\text{,MC}}}
\newcommand*{\mucopRQMC}{\hat{\mu}_{\ngen}^{C}}
\newcommand*{\muGMMNMC}{\hat{\mu}_{\ngen}^{\text{GMMN,MC}}}
\newcommand*{\muGMMNRQMC}{\hat{\mu}_{\ngen}^{\text{GMMN}}}
\newcommand*{\muAGMMNMC}{\hat{\mu}_{\ngen}^{\text{AGMMN,MC}}}
\newcommand*{\muAGMMNRQMC}{\hat{\mu}_{\ngen}^{\text{AGMMN}}}
\newcommand*{\mudot}{\hat{\mu}_{\ngen}^{\,\bigcdot}}
\newcommand*{\muMC}{\hat{\mu}_{\ngen}^{\text{MC}}}

\makeatletter
\newcommand*\bigcdot{\mathpalette\bigcdot@{.5}}
\newcommand*\bigcdot@[2]{\mathbin{\vcenter{\hbox{\scalebox{#2}{$\m@th#1\bullet$}}}}}
\makeatother

\begin{document}
\thispagestyle{plain}
\begin{center}
	\sffamily
	{\bfseries\LARGE Adaptive Generative Moment Matching Networks for Improved Learning of Dependence Structures\par}
	\bigskip\smallskip
	{\Large \ifunblinded Marius Hofert\footnote{Department of Statistics and Actuarial Science, The University of
				Hong Kong,
				\href{mailto:mhofert@hku.hk}{\nolinkurl{mhofert@hku.hk}}.},
			\Large Gan Yao\footnote{Department of Statistics and Actuarial Science, The University of
				Hong Kong,
				\href{mailto:ganyao@connect.hku.hk}{\nolinkurl{ganyao@connect.hku.hk}}.}\fi
		\par
		\bigskip
		\today\par}
\end{center}
\par\smallskip
\begin{abstract}
	An adaptive bandwidth selection procedure for the mixture kernel in the maximum
	mean discrepancy (MMD) for fitting generative moment matching networks (GMMNs) is
	introduced, and improved learning of copula random number generators is
	demonstrated. Based on the relative error of the training loss, the number of
	kernels is increased during training; additionally, the relative error of the
	validation loss is used as an early stopping criterion. While training time remains
	similar, adaptively training GMMNs (AGMMNs) significantly increases training
	performance, which is shown based on validation MMD trajectories, samples and
	validation MMD values. Superiority of AGMMNs over GMMNs and parametric copula
	models is also demonstrated in terms of three applications. First, convergence
	rates of estimators based on quasi-random versus pseudo-random samples from copulas
	are investigated in dimensions as large as 100 for the first time. Second,
	replicated validation MMDs, as well as Monte Carlo and quasi-Monte Carlo
	applications demonstrate the improved training of AGMMNs for a copula model implied
	by the 50 constituents of the S\&P~500 index after deGARCHing. Last, both the
	latter dataset and 50 constituents of the FTSE~100 are used to demonstrate that the
	improved training of AGMMNs indeed translates to an improved model prediction.
\end{abstract}
\minisec{Keywords}
Bandwidth selection,
copula,
GMMN,
high dimensions,
maximum mean discrepancy,
quasi-random sampling.
\minisec{MSC2010}
62H99, 65C60, 60E05, 00A72, 65C10. %

\section{Introduction}
Fitting and then simulating from realistic copulas is central to predictive
modeling in statistics, finance, insurance and quantitative risk management. With
focus on high dimensions, we present an adaptive training strategy of certain
neural networks for generative modeling of an arbitrary copula and demonstrate its
effectiveness in several applications.

Copulas are $d$-dimensional distribution functions with standard uniform univariate
margins that uniquely characterize the dependence structure of a $d$-dimensional
distribution function with continuous margins (\cite{sklar1959}).
Although various parametric copula models (\cite{hofertvrins2013},
\cite{hoferthuserprasad2018}, \cite{hintzhofertlemieux2020a}, \cite{hofert2021},
\cite{hofertziegel2021}, \cite{herrmannhofertsadr2023}, \cite{gumbel1960b},
\cite{gumbel1961}, \cite{galambos1975}, \cite{abdousfougeresghoudi2005},
\cite{hashorva2005}) and sampling methods (\cite{whelan2004}, \cite{mcneil2008},
\cite{hofert2011a}, \cite{hofert2012b}, \cite{grothehofert2015},
\cite{arbenzcambouhofertlemieuxtaniguchi2018}, \cite{hintzhofertlemieux2022c},
\cite{dombryengelkeoesting2016}) exist, classical copula modeling faces challenges
such as slow inference, model inadequacy and numerically challenging
implementations in high dimensions. On the other hand, generative neural networks
are trained to directly transform a sample of a known input distribution
$F_{\bm{Z}}$ (the \emph{prior} distribution) to a sample of the \emph{target
	copula} $C$ (\cite{hofertprasadzhu2021a}, \cite{jankeghanmisteike2021},
\cite{hofertprasadzhu2022a}). They thus enable purely data-driven copula modeling,
in a flexible and numerically robust manner.

We consider a multi-layer perceptron (MLP) neural network trained with the loss
function being the maximum mean discrepancy (MMD) of
\cite{grettonborgwardtraschschoelkopfsmola2007}, a measure of discrepancy between
model generated and observed data (\cite{liswerskyzemel2015},
\cite{lichangchengyangpoczos2017}, \cite{hofertprasadzhu2021a},
\cite{hofertprasadzhu2022a}, \cite{hofertprasadzhu2023a},
\cite{hofertprasadzhu2023b}). The MMD depends on a kernel $k(\cdot,\cdot): \IR^d
	\times \IR^d \rightarrow \IR$ as a measure of similarity between two data points. A
choice of kernel successful in some applications is the mixture radial basis
function (RBF) kernel (\cite{liswerskyzemel2015}). The superiority of RBF kernels
and their mixtures lies in them being \emph{characteristic} %
(\cite{fukumizugrettonsunscholkopf2007},
\cite{sriperumbudurgrettonfukumizuscholkopflanckriet2010}), which gives the
theoretical guarantee that the MMD between two Borel probability measures $\P$ and
$\mathbb{Q}$ is zero if and only if $\P = \mathbb{Q}$. Training an MLP with the MMD
as loss function results in a \emph{generative moment matching network (GMMN)}
(\cite{liswerskyzemel2015}, \cite{dziugaiteroyghahramani2015}).

To compute the MMD between a $d$-dimensional training sample
$X=(\bm{X}_1\T,\dots,\bm{X}_{\ntrn}\T)\T\in\IR^{\ntrn\times d}$ of size $\ntrn$ and
a $d$-dimensional neural network generated sample $Y=(\bm{Y}_1\T,\dots,$
$\bm{Y}_{\ngen}\T)\T\in\IR^{\ngen\times d}$ of size $\ngen$ with a mixture of $\nkrn$ RBF
kernels as kernel, we need to select the bandwidths $\bm{h}=(h_1,\dots,h_{\nkrn})$
the underlying kernels depend on. Early references on the MMD usually consider
$\nkrn = 1$ and use the median heuristic of
\cite{sriperumbudurfukumizugrettonlanckrietscholkopf2009}, which chooses the single
bandwidth parameter as the median of all pairwise distances in the
$(2\ntrn,d)$-matrix of stacked samples of $X$ and $Y$. Although possessing
theoretical merits, \cite{ramdasreddipoczossinghwasserman2015} and
\cite{garreaujitkrittumkanagawa2018} noted that the median heuristic for a single
bandwidth only works in a limited amount of setups. \cite{liswerskyzemel2015} thus
considered mixtures of kernels.
Alternatively, optimizing certain objective functions to obtain optimal bandwidths
was explored in the context of multiple kernel learning by
\cite{gonenalpaydin2011}.
\cite{grettonsriperumbudursejdinovicstrathmannbalakrishnanPontilfukumizu2012}
propose to optimize the power of an MMD two-sample test by maximizing the ratio of
an MMD estimate linear in sample size %
over its estimated variance.
\cite{sutherlandtungstrathmannderamdassmolagretton2017} extend this idea to an MMD
estimate quadratic in sample size, which can achieve a higher power.
Yet another idea is to bypass the optimization of bandwidths and to learn a
\emph{deep kernel} of the form $k(\bm{x}, \bm{y}) = k_0(\phi(\bm{x}),
\phi(\bm{y}))$ that can adapt to a chosen bandwidth, where $k_0$ is a kernel and
$\phi$ a deep feature map. \cite{lichangchengyangpoczos2017} propose MMD GANs,
where they combine GMMNs with deep kernel learning and adversarially train a GMMN
with an MMD that uses a deep kernel. Adversarial training of generative models is
notoriously difficult, though, which contrasts the attractive simplicity of GMMNs.

For training random number generators from copulas, \cite{hofertprasadzhu2021a},
\cite{hofertprasadzhu2022a}, \cite{hofertprasadzhu2023a} and \cite{hofertprasadzhu2023b}
used $\nkrn=6$ and $\bm{h}=(0.001,0.01,0.15,0.25,0.50,$ $0.75)$.
When learning random
number generators in high dimensions, hard-coded bandwidths parameters are
typically not adequate anymore (see also our results later). This limitation has
partly already been demonstrated in \cite{hofertprasadzhu2021a}, where the
performance of GMMNs as quasi-random number generators was shown to decrease
with increasing dimensions up to 10. We write ``partly'' as it remained unclear
whether this decrease in variance reduction can be attributed to a lack of
massive training data to learn a higher-dimensional distribution properly or a
decrease in the variance-reduction capability of the underlying low-discrepancy
point set; as we will see later, when investigating much larger dimensions, the
latter indeed plays a role.

In Section~\ref{sec:methods}, we introduce a training approach that adaptively
selects the bandwidths of the mixture RBF kernel in order for the MMD loss to
better adapt to the characteristics of the training data than hard-coded values
can, while avoiding having to solve a more complex optimization problem or
having to train
another neural network. As we will demonstrate in the sections
thereafter, the resulting adaptively trained GMMNs, or \emph{adaptive GMMNs
  (AGMMNs)} in short, result in a notably better performance, especially in
higher dimensions. In particular, in Section~\ref{sec:assess_fit}, we assess
AGMMNs in terms of the quality of the generated samples. Section~\ref{sec:rqmc}
studies the performance of AGMMNs when the prior is a low-discrepancy point set,
which is important in randomized quasi-Monte Carlo (RQMC) applications in the
realm of predictive modeling. Then, in Sections~\ref{sec:simulation} and
\ref{sec:prediction}, we showcase AGMMNs and other models on a synthetic dataset
and two real-world datasets, respectively, and demonstrate that AGMMNs provide
improved Monte Carlo (MC) estimators. Finally, Section~\ref{sec:concl}
summarizes our contributions.

\section{Adaptive generative moment matching networks}\label{sec:methods}
For Borel probability measures $\P, \mathbb{Q}$, the squared MMD is defined by
$\MMD^2(\mathcal{F}, \P, \mathbb{Q}) = \bigl(\sup_{f \in \mathcal{F}} ($
$\E_{\P}(f) - \E_{\mathbb{Q}}(f)) \bigr)^2 = \lVert \mu_{\P} -
\mu_{\mathbb{Q}}\rVert_{\mathcal{H}}^2 $, where $\mathcal{F}$ is a unit ball in
a reproducing kernel Hilbert space (RKHS) $\mathcal{H}$ and $\mu_{\P}$ and
$\mu_{\mathbb{Q}}$ are mean embeddings of $\P$ and $\mathbb{Q}$ in
$\mathcal{H}$, respectively, such that
$\E_{\P}(f) = \langle f, \mu_{\P} \rangle_{\mathcal{H}}$ and
$\E_{\mathbb{Q}}(f) = \langle f, \mu_{\mathbb{Q}} \rangle_{\mathcal{H}}$. The
sample version of the $\MMD^2$ based on two datasets
$U=(\bm{U}_1\T,\dots,\bm{U}_{\ntrn}\T)\T\in[0,1]^{\ntrn\times d}$,
$Y=(\bm{Y}_1\T,\dots,\bm{Y}_{\ntrn}\T)\T\in[0,1]^{\ntrn\times d}$ and kernel
function $k(\cdot,\cdot): \IR^d \times \IR^d \rightarrow \IR$
is %
$\MMD(U,Y)^2=\frac{1}{\ntrn^2}
\sum_{i_1=1}^{\ntrn}\sum_{i_2=1}^{\ntrn}(k(\bm{U}_{i_1},\bm{U}_{i_2})-
2k(\bm{U}_{i_1},\bm{Y}_{i_2}) + k(\bm{Y}_{i_1},\bm{Y}_{i_2}))$. To allow for
better discrimination, the (single) kernel $k$ is often replaced by a mixture, a
popular choice being the \emph{mixture RBF kernel}
\begin{align}
  k(\bm{u},\bm{y};\bm{h}) =\sum_{l=1}^{\nkrn} k(\bm{u},\bm{y};h_l),\label{eq:kernel_mixture}%
\end{align}
where $\nkrn$ denotes the number of kernels and
\begin{align}
  k(\bm{u},\bm{y};h) = e^{-\frac{1}{2h^2}\lVert \bm{u}-\bm{y}\rVert_2^2}, \label{eq:kernel_rbf}
\end{align} is
the \emph{RBF kernel} with bandwidth $h>0$. The resulting sample version of the $\MMD$ is
then
\begin{align}
  \MMD(U,Y;\bm{h})=\sqrt{\frac{1}{\ntrn^2} \sum_{i_1=1}^{\ntrn}\sum_{i_2=1}^{\ntrn}(k(\bm{U}_{i_1},\bm{U}_{i_2};\bm{h})- 2k(\bm{U}_{i_1},\bm{Y}_{i_2};\bm{h}) + k(\bm{Y}_{i_1},\bm{Y}_{i_2};\bm{h}))}.\label{eq:MMD}
\end{align}
We also compute the \emph{average MMD}
\begin{align}
  \overline{\MMD}_{\nrep}(\{U_i\}_{i=1}^{\nrep}, \{Y_i\}_{i=1}^{\nrep};\bm{h})=\frac{1}{\nrep} \sum_{i=1}^{\nrep}\MMD(U_i,Y_i;\bm{h}),\label{eq:mean:MMD}
\end{align}
of the $\MMD$ loss function based on samples $U_i$ and $Y_i$, $i=1,\dots,n$,
namely for the training loss when performing batch optimization, as well as for
determining the validation loss (see later for details).

We see from~\eqref{eq:kernel_rbf} that bandwidths should be of the same order as
$\lVert \bm{u}-\bm{y}\rVert_2$ since, otherwise, the resulting MMD values are
too close to $0$, which may reduce the discrimination effectiveness of the MMD
as a loss function for training generative models. In light of this and inspired
by the median heuristic, we propose to select bandwidths as empirical quantiles
of pairwise distances in the training data; these empirical quantiles only have
to be determined once. We consider pairwise distances of training samples only
(instead of stacked training samples or generated samples) due to the fact that
the distribution of the model-generated samples quickly approaches that of
training samples after a few training epochs. More importantly, rather than
using one set of bandwidths during the entire course of training, we suggest to
adaptively update all bandwidths during training, where, on each update, the
number of kernels $\nkrn$ is increased. We call epochs of training sharing the
same (number of) kernels and bandwidths a training \emph{phase}. As we will see,
training in phases can make the model converge faster, and it can lead to the
convergence to a better optimum. As already mentioned, we call a GMMN trained
with this adaptive kernel bandwidth selection procedure an \emph{adaptive GMMN
  (AGMMN)}.

GMMNs are similar to GAN-like models, the class of models that resemble generative
adversarial networks (GANs) (\cite{goodfellow2014}), where the MMD in the former
can be viewed as an equivalent to the critic network typically found in the latter.
A complex critic usually fails to provide useful gradients when the generated
distribution is still far from the target distribution. However, as the model
improves, a simple critic becomes insufficient for distinguishing finer
discrepancies, and a complex critic is needed; see
\cite{arbelsutherlandbinkowskigretton2018} in the context of GAN-like models. The
intuition behind our proposed AGMMN is thus to start training with a critic that is
relatively simple and then gradually increase its complexity by increasing the
number of kernels $\nkrn$ as training proceeds.

Unlike the usual training of GMMNs, updating bandwidths during training implies
that the loss function itself is modified during training, which makes it
difficult to determine when to stop based on the training loss alone. As a
solution, we evaluate model performance with the \emph{validation MMD} that is
the MMD~\eqref{eq:MMD} with the RBF kernel~\eqref{eq:kernel_rbf} and bandwidth
$\bm{h}$ fixed at
\begin{align}
  \hval = (0.05,0.1,0.2,0.3,0.4,0.5,0.6,0.7,0.8,0.9,0.95).\label{eq:hval}
\end{align}
During training, we compute and monitor its average
$\overline{\MMD}_{\nrep}(\{U_i\}_{i=1}^{\nrep}, \{Y_i\}_{i=1}^{\nrep};\hval)$ as
in~\eqref{eq:mean:MMD} and consider it as the \emph{validation loss}.  Since
$\hval$ is fixed, the validation loss
$\overline{\MMD}_{\nrep}(\{U_i\}_{i=1}^{\nrep},$ $\{Y_i\}_{i=1}^{\nrep};\hval)$ is
then comparable across different training epochs. Furthermore, it can be used to
determine early stopping of the training procedure, which is important for
practical applications. In particular, if the training loss stabilizes, we
either stop training entirely (if we are already in the final training
phase) %
or, if not, update the bandwidths to enter the next training phase (see
Algorithm~\ref{alg:AGMMN} Step~\ref{alg:stop:entirely:or:update:bandwidths}).
And if the validation loss stabilizes, we stop updating bandwidths (see
Algorithm~\ref{alg:AGMMN} Step~\ref{alg:stop:updating}), thus do not consider
further training phases.

Let us now turn to the adaptive selection of bandwidths. Based on the training sample
$U \in [0,1]^{\ntrn \times d}$, let $\mathcal{D}(U) = \{\lVert\bm{U}_{i_1} - \bm{U}_{i_2}\rVert\}_{1\le i_1 < i_2
  \le \ntrn}$ be the set of pairwise distances with empirical distribution function
\begin{align*}
  \hat{F}_{\mathcal{D}(U)}(x) = \frac{2}{n(n-1)}\sum_{1 \le i_1 < i_2 \le \ntrn}\I_{\{\lVert \bm{U}_{i_1} - \bm{U}_{i_2}\rVert \le x \}},\quad x\in\IR,
\end{align*}
and corresponding empirical quantile function
$\hat{F}^{-1}_{\mathcal{D}(U)}(y) = \inf \{x \in \IR:
\hat{F}_{\mathcal{D}(U)}(x)\ge y\}$, $y\in(0,1)$. Let $\nepo\in\IN$
  denote the maximal number of training epochs and $\nphs\in\IN$ the maximal
  number of training phases considered for training an AGMMN.
  Let $(n_{\krn,1},\dots,n_{\krn,\nphs})$ be the vector of increasing
  number of kernels used over the course of training, where $n_{\krn,k}$ stands for
  the number of kernels used in training phase $k$.  Now let
  $(\bm{p}_k)_{k = 1}^{\nphs}$ with
  $\bm{p}_k = (p_{k,1}, \dots, p_{k,n_{\krn,k}})\in[0,1]^{n_{\krn,k}}$,
  $k = 1, \ldots, \nphs$, denote a sequence of vectors of probabilities
  determining the bandwidths $\bm{h}_k$ used in training phase $k$ via
  \begin{align}
    \bm{h}_k=\hat{F}_{\mathcal{D}(U)}^{-1}(\bm{p}_k),\quad k = 1, \ldots, \nphs.\label{eq:bandwidth:vecs}
  \end{align}
  Due to the adaptive choice of bandwidths entering the
  kernel of the MMD loss function in training phase $k$, we also adjust the
  learning rate $\gamma_k>0$ of the $k$th epoch (to be detailed later).
  Once a new training phase is entered in epoch $t$, training takes place at
  least over the \emph{patience} $r_t$ many epochs (also to be detailed later). After
  that, on every new epoch in the current training phase, the training
  loss is checked for convergence, which is determined by the relative error
  of the training loss failing to improve by more than a threshold
  $\Delta_{\text{trn}}\ge 0$ over the course of the past $r_t$ epochs.
  Once per training phase, also the validation loss is checked for convergence,
  which is determined similarly as for the training loss but based on the error
  threshold $\Delta_{\text{val}}\ge 0$. If the validation loss converged in the
  current training phase, training may terminate if the training loss converged.
  If the latter did not converge, end this training phase, update the
  vector of bandwidths according to~\eqref{eq:bandwidth:vecs} and continue
  training in the next training phase.
  The following algorithm summarizes our proposed training procedure of AGMMNs
  and contains more details, e.g.\ concerning non-convergence if $\nepo$ or
  $\nphs$ are chosen too small for the problem at hand. Exact parameter choices
  are addressed thereafter.

\begin{algorithm}[AGMMN training]\label{alg:AGMMN}
  Let $U=(\bm{U}_1\T,\dots,\bm{U}_{\ntrn}\T)\T\in [0,1]^{\ntrn \times d}$ be the
  training sample of size $\ntrn\in\IN$, $\nbat \in\{1, \dots, \ntrn\}$ the
  batch size, $\nbts=\lceil \ntrn/\nbat \rceil$ the number of batches (the last
  batch may have a size smaller than $\nbat$), %
  $\nrep\in\IN$ the number of datasets for determining the validation loss,
  $\ndat\in\IN$ the sample size of each of the $\nrep$ datasets for determining
  the validation loss, $\hval$ the validation bandwidths as in~\eqref{eq:hval},
  $\nepo\in\IN$ the maximal number of training epochs,
  $r_1,\dots,r_{\nepo}\in\IN$ the patience parameters for each training epoch
  $\nphs\in\IN$ the maximal number of training phases,
  $(\bm{p}_k)_{k=1}^{\nphs}$ probability vectors with
  $\bm{p}_k = (p_{k,1}, \dots, p_{k,n_{\krn,k}})\in[0,1]^{n_{\krn,k}}$,
  $k=1, \ldots, \nphs$, $\gamma_1,\dots,\gamma_{\nphs}>0$ the learning rates for
  each training phase, and $\Delta_{\text{trn}} \geq 0$ and
  $\Delta_{\text{val}} \geq 0$ the relative error thresholds for determining
  convergence of the training and validation loss, respectively.
  \begin{enumerate}
  \item \textbf{Initialization:} Initialize the MLP weights and biases of each layer with input dimension
    $d_{\text{in}}$ via the uniform distribution
    $\U(-\frac{1}{\sqrt{d_{\text{in}}}}, \frac{1}{\sqrt{d_{\text{in}}}})$; this is also
    the default of \texttt{PyTorch}. %
    For training phase $1$, set the phase counter $k\leftarrow 1$, compute
    the bandwidth vector
    $\bm{h}_k = \hat{F}_{\mathcal{D}(U)}^{-1}(\bm{p}_k)$ as
    in~\eqref{eq:bandwidth:vecs}, set the epoch of last bandwidth update
    $t_{\text{up}}\leftarrow 1$, and set the logical variable indicating when to
    stop considering updating bandwidths (so when to stop entering new training
    phases) $\STOP_{\text{up}}\leftarrow\text{FALSE}$.
  \item\label{alg_step:updates} For epoch $t=1,\dots,\nepo$, do: %
    \begin{enumerate}
    \item \textbf{Training in epoch $t$:} Generate a sample
      $Y=(\bm{Y}_1\T,\dots,\bm{Y}_{\ntrn}\T)\T\in [0,1]^{\ntrn \times d}$ from the
      MLP performing a forward pass of a sample of the same size from the prior
      distribution. Randomly shuffle the rows of $U$. Let $U_b,Y_b$ denote the
      $b$th of the $\nbts$ mini-batches of $U,Y$, respectively.  For each
      $b=1,\dots,\nbts$, compute the training loss $\MMD(U_b,Y_b;\bm{h}_k)$ and take
      a gradient step with learning rate $\gamma_k$.
    \item \textbf{Determine the training loss in epoch $t$:}
      Compute the average training loss in epoch $t$ as
      $L_{\text{trn}}(t)=\overline{\MMD}_{\nbts}(\{U_b\}_{b=1}^{\nbts},\{Y_b\}_{b=1}^{\nbts};\bm{h}_k)$.
    \item \textbf{Determine the validation loss in epoch $t$:}
      For $i=1,\dots,\nrep$, do:
      \begin{enumerate}
      \item Determine a sample $\tilde{U}_i$ of size $\ndat\times d$ by randomly
        sampling rows of the training data $U$ with replacement.
      \item Generate a sample $\tilde{Y}_i$ of size $\ndat\times d$ from the MLP by
        performing a forward pass of a sample of the same size from the prior
        distribution.
      \end{enumerate}
      Compute the validation loss
      $L_{\text{val}}(t)=\overline{\MMD}_{\nrep}(\{\tilde{U}_i\}_{i=1}^{\nrep},\{\tilde{Y}_i\}_{i=1}^{\nrep};\hval)$
      as in~\eqref{eq:mean:MMD} based on~\eqref{eq:hval}.
    \item\label{alg:stop:updating} \textbf{Determine convergence of the validation loss:}
    If $t = t_{\text{up}} + r_t$ %
    and $1 - \frac{L_{\text{val}}(t-s)}{L_{\text{val}}(t_{\text{up}})} \leq
    \Delta_{\text{val}}$ for all $s = 0,1, 2, \dots, r_t-1$, set $\STOP_{\text{up}}$ $\leftarrow$ TRUE
    (indicating that training finishes successfully once the training loss passes its convergence check).
    \item\label{alg:stop:entirely:or:update:bandwidths} \textbf{Determine convergence of the training loss:}
      If $t\ge t_{\text{up}}+r_t$
      and $1 - \frac{L_{\text{trn}}(t-s)}{L_{\text{trn}}(t-r_t)} \leq \Delta_{\text{trn}}$
      for all $s = 0,1,2,\dots,r_t-1$, do:
      \begin{enumerate}
      \item If $\STOP_{\text{up}}$ $=$ TRUE (indicating no further bandwidth update shall
        take place; equivalently, no further training phase shall be entered),
        then return the trained AGMMN (as both training and validation loss have converged).
      \item
        Otherwise, %
        increase the phase counter $k\leftarrow k+1$.
        \begin{enumerate}
        \item If $k\le \nphs$, compute the updated bandwidth vector $\bm{h}_k$
          based on $\bm{p}_k$ as in~\eqref{eq:bandwidth:vecs} and record the
          current epoch as epoch of last bandwidth update by setting
          $t_{\text{up}}\leftarrow t$.
        \item If $k>\nphs$,
          abort training (as only the training but not the validation loss has
          converged yet, and no more training phases are allowed to be run). %
        \end{enumerate}
      \end{enumerate}
    \end{enumerate}
  \item If $t=\nepo$, abort training (as the validation loss has never converged
    or it has, but the training loss has never converged).
  \end{enumerate}
\end{algorithm}

We now address the network architecture and parameter choices. We follow
\cite{hofertprasadzhu2021a} and utilize the MLP with ReLU activation functions
for hidden layers and sigmoid for the output layer. In
Sections~\ref{sec:assess_fit} to \ref{sec:simulation} we use a single hidden
layer with 300 neurons, as in \cite{hofertprasadzhu2021a}. In
Section~\ref{sec:prediction} we also investigate a deeper architecture with two
hidden layers of sizes 2000 and 500, respectively, and a wide one with 10\,000
neurons in a single hidden layer. As prior distribution $F_{\bm{Z}}$, we use
$\N_{\dpri}(\bm{0},I_{\dpri})$ with $\dpri=d$ throughout, except for
Section~\ref{sec:rqmc}, where we also experiment with $\dpri < d$.  Parameters
that remain fixed in Sections~\ref{sec:assess_fit} to \ref{sec:prediction} are
specified as follows; application-specific setups (e.g.\ types of copulas,
strength of dependencies, dimensions $d$) and parameters (e.g.\ training sample
size $\ntrn$, batch size $\nbat$) will be detailed in the respective sections.
For the patience in epoch $t=1,\dots,\nepo$, we found through experimentation
that
\begin{align}
  r_t = \bigl\lfloor\min\bigl\{50, \max\bigl\{20, \tfrac{3}{8}(t - 20)\bigr\}\bigr\}\bigr\rfloor = \begin{cases}
    20,                                                 & t \leq 20,       \\
    \bigl\lfloor 20 + \tfrac{3}{8}(t - 20)\bigr\rfloor, & 20 < t \leq 100, \\
    50,                                                 & t > 100.
  \end{cases}\label{eq:patience:choice}
\end{align}
ensures that the model is sufficiently optimized with respect to each set of
bandwidths before updating to the next such set, which then results in a faster
and more robust convergence behavior of the training procedure. The maximal
number of training epochs is set to be $\nepo = 800$ and the maximal number of
training phases is chosen as $\nphs=4$, which is sufficient for early stopping
in all examples run. The vector of number of kernels over all training phases is
$(n_{\krn,1},\dots,n_{\krn,\nphs})=(6, 12, 24, 48)$.  Since we found by
experimentation that a bandwidth vector denser around the left tail of
$\hat{F}_{\mathcal{D}(U)}$ improves learning, we specify
\begin{align*}
  p_{k,l} = 0.95 \times 2^{-9\frac{n_{\krn,k}-l}{n_{\krn,k}}},\quad k = 1, \ldots, \nphs,\ l=1,\dots,n_{\krn,k},
\end{align*}
which is equidistant in log-scale, ending in 0.95 and denser near $0$ with every
phase. An intuition behind this choice is that smaller bandwidths can help
capture finer differences between the target- and model-generated
distributions. For optimization, we use the Adam optimizer of \cite{kingma2015}
with learning rates $\gamma_k = 0.001\cdot 5^{-(k-1)}$, $k=1,\dots,\nphs$, in
order to account for the gradient inflation partially induced by the increase in
the number of kernels over different training phases and to improve the
convergence of the training loss in later phases.  The number of datasets
$\nrep$ generated for computing the validation loss during training is chosen as
$\nrep=50$ and the size of each dataset is set to be $\ndat=3000$. Finally, the
relative convergence tolerances where chosen as $\Delta_{\text{trn}}=0$ and
$\Delta_{\text{val}}=0.05$ throughout.

MLP architectures are smooth almost everywhere, which helps to preserve the low
discrepancy of RQMC point sets after they are passed through the neural network.
This suggests the use of AGMMN RQMC estimators for better performance over
AGMMN-based Monte Carlo (MC) estimators, even when the AGMMN is trained on
pseudo-random samples. Quasi-random sampling from AGMMNs then follows the same
steps as for GMMNs in \cite{hofertprasadzhu2021a}, which are summarized in the
following algorithm.
\begin{algorithm}[Quasi-random sampling from AGMMNs]\label{alg:QRNG:AGMMN}
  \begin{enumerate}
  \item Let $f_{\hat{\bm{\theta}}}: \IR^{\dpri} \to [0,1]^d$ be an AGMMN model
    trained on pseudo-observations of a target copula $C$. Let $\ngen \in \IN$
    be the size of the sample to be generated from the target copula.
  \item Generate an RQMC point set $\tilde{P}_{\ngen} = \{\tilde{\bm{v}}_1, \dots,
    \tilde{\bm{v}}_{\ngen}\}$, $\tilde{\bm{v}}_i \in [0, 1]^{\dpri}$, $i = 1, \dots,
    \ngen$, and compute $\bm{Z}_i = F^{-1}_{\bm{Z}}(\tilde{\bm{v}}_i)$, $i = 1, \dots,
    \ngen$. \item\label{alg:QRNG:AGMMN:postprocess} Pass $\bm{Z}_i$ through the AGMMN to obtain $\bm{Y}_i =
    f_{\hat{\bm{\theta}}}(\bm{Z}_i)$, $i = 1, \dots, \ngen$, then compute and return
    the pseudo-observations $\bm{U}_1,\dots,\bm{U}_{\ngen}$ of
    $\bm{Y}_1,\dots,\bm{Y}_{\ngen}$ via
    \begin{align*}
      U_{i,j}=\frac{1}{\ngen+1}\sum_{k=1}^{\ngen}\I_{\{Y_{k,j}\le Y_{i,j}\}},\quad i=1,\dots,\ngen,\ j=1,\dots,d
    \end{align*}
    as a quasi-random sample from the target copula $C$.
  \end{enumerate}
\end{algorithm}
For all applications involving quasi-random sampling, we use digital-shifted
Sobol' sequences as RQMC point sets (\cite{lemieux2009}).  Finally, in
Sections~\ref{sec:assess_fit} to \ref{sec:prediction} we use GMMNs trained with
fixed bandwidth vector (chosen as the default of the \R\ package \texttt{gnn}
developed based on \cite{hofertprasadzhu2021a}) as a benchmark %
to compare our AGMMN against. The acronym ``GMMN'' is therefore specifically used to refer to
GMMNs trained with this fixed bandwidth vector.

\section{Assessing AGMMN samples}\label{sec:assess_fit}
\subsection{Model architecture and training setup}\label{subsec:setup}
All experiments in this section share the following training setup. As training
sample size, we use $\ntrn = 60\,000$ and conduct training in mini-batches of size
$\nbat = 3000$.

As target copulas from which training data are generated, we consider different
families with varying strengths of dependence in the large dimension $d =
100$. The copulas include a Clayton copula with pairwise Kendall's tau
$\tau\in\{0.25, 0.5\}$, a Gumbel copula with pairwise Kendall's tau
$\tau\in\{0.25, 0.5\}$, an exchangeable normal copula with exchangeable
dispersion correlation matrix implied by a pairwise Kendall's tau
$\tau\in\{0.25, 0.5\}$, and an exchangeable Student's $t$ copula with $\nu=4$
degrees of freedom and the same exchangeable dispersion correlation matrices as
for the normal copulas.  In addition, two non-exchangeable copula models are
considered. One is a two-level nested Gumbel copula with first-level Kendall's
tau $\tau=0.3$ and ten second-level 10-dimensional copulas having
parameters %
linearly interpolated between $\tau^{-1}(0.3)$ and $\tau^{-1}(0.6)$; note that
for Gumbel copulas with parameter $\theta$, one has
$\tau(\theta)=\frac{\theta-1}{\theta}$ so that
$\tau^{-1}(\tilde{\tau})=(1-\tilde{\tau})^{-1}$.  The other non-exchangeable
model is a D-Vine copula, where, at each tree level, edgewise copula families
are randomly sampled from a set including Gaussian, $t$ with degrees of freedom
randomly sampled from $[4, 10]$, Clayton, survival Clayton, Gumbel and survival
Gumbel, and the corresponding dependence parameters decay with tree level and
are perturbed by random noises.

\subsection{Effect of adaptively updating the bandwidths on the validation loss}
To understand how the adaptively updated bandwidths in Algorithm~\ref{alg:AGMMN}
help the learning process, we compare AGMMN validation loss trajectories with
those of models that have the same architecture and are also trained with
Algorithm~\ref{alg:AGMMN}, except that the number of kernels $\nkrn$ are kept
fixed and we train over $\nepo = 800$ epochs without early stopping.  We refer
to such a model as $\AGMMN_{\nkrn}$ in what follows. The most notable difference
between a GMMN and $\AGMMN_{\nkrn}$ is how the bandwidths are determined (GMMN:
six kernels and fixed bandwidths; AGMMN: $\nkrn$ kernels and bandwidths
determined from the training data). We train $\AGMMN_{\nkrn}$ for
$\nkrn\in\{6, 12, 24, 48\}$.

\begin{figure}[htbp]
  \includegraphics[width=0.49\textwidth]{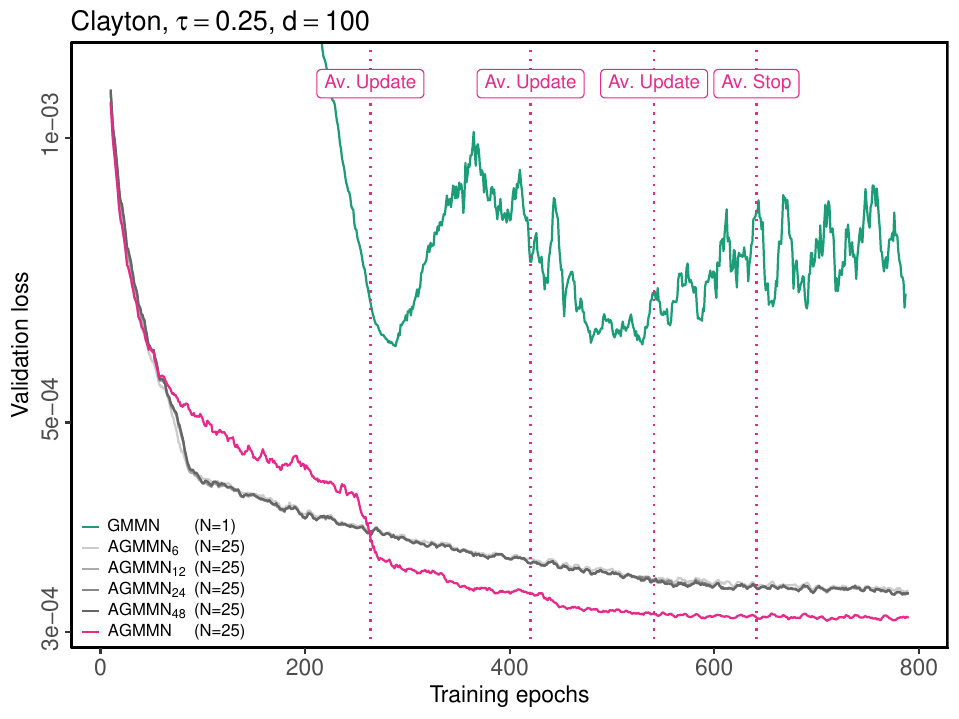}\hfill
  \includegraphics[width=0.49\textwidth]{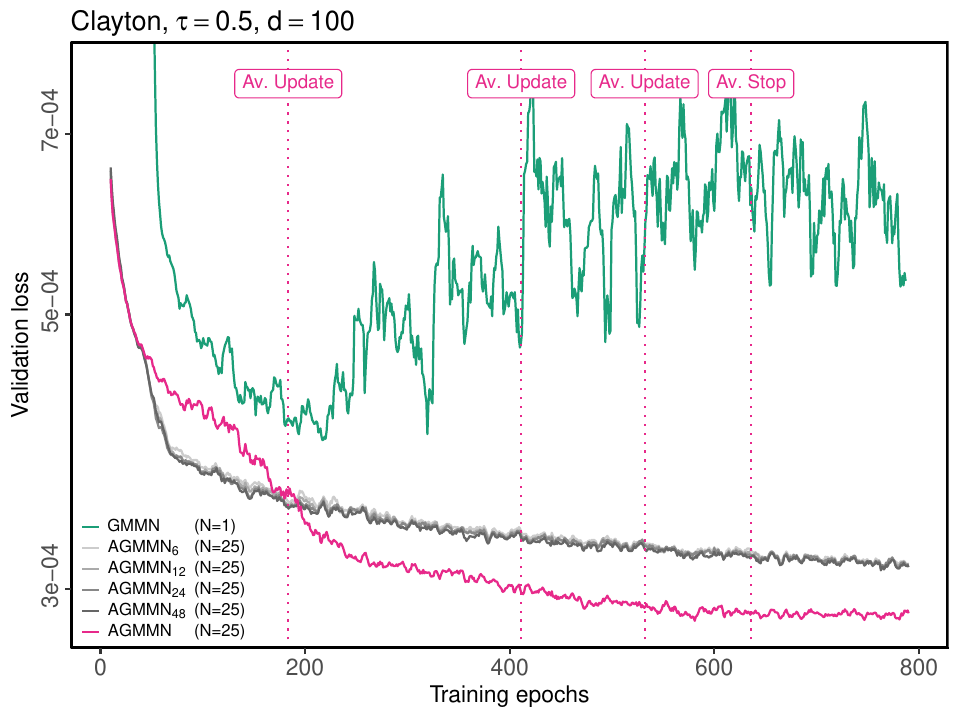}\\[2mm]
  \includegraphics[width=0.49\textwidth]{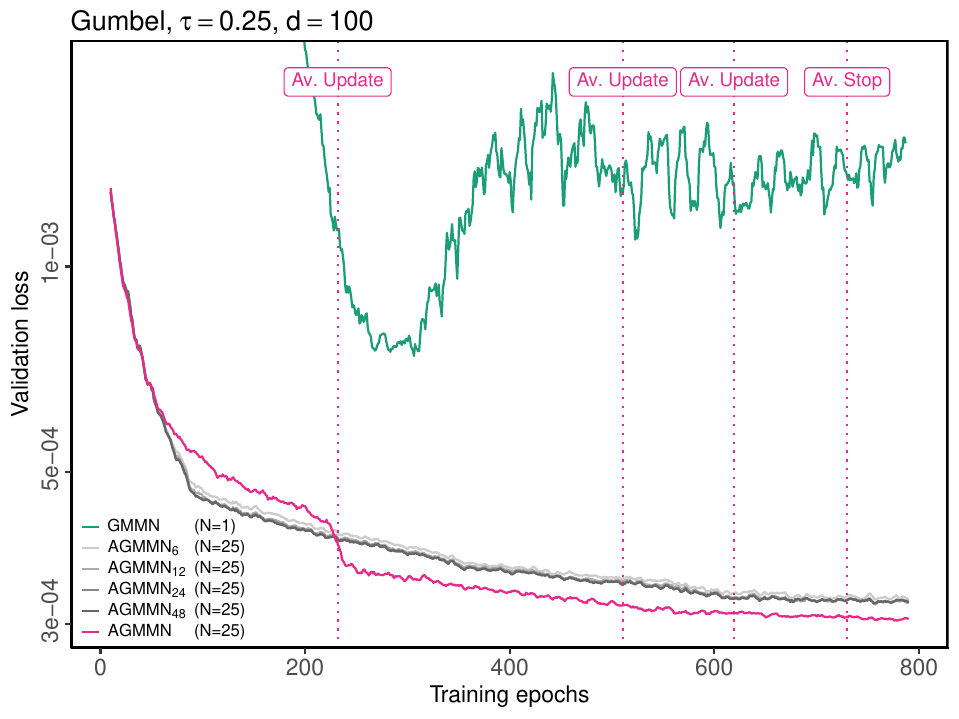}\hfill
  \includegraphics[width=0.49\textwidth]{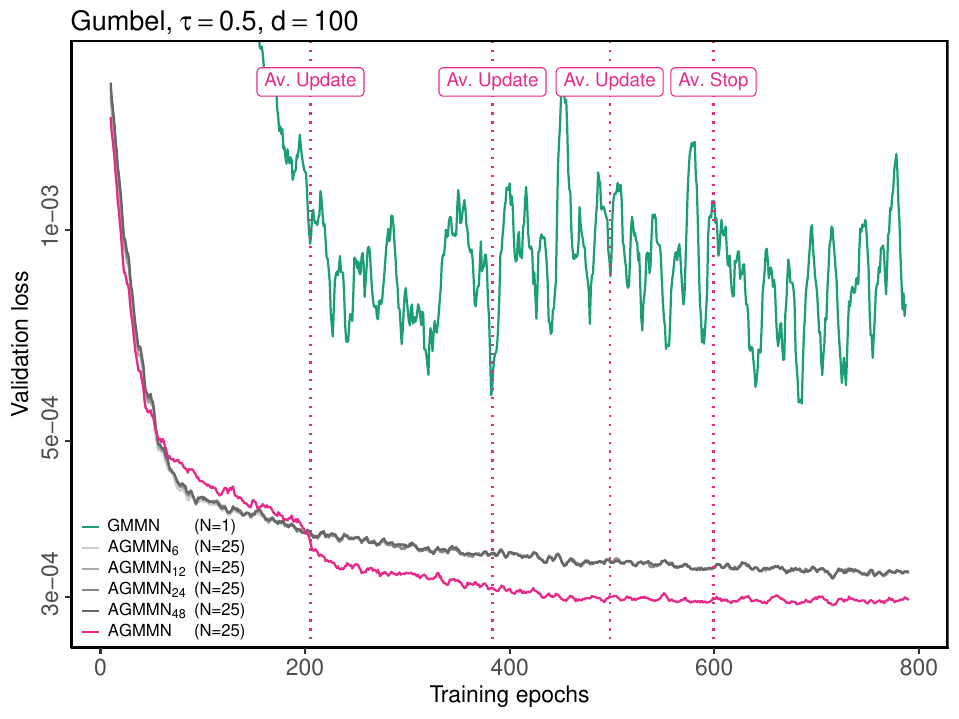}
  \caption{Validation loss trajectory of a GMMN, %
      median validation loss trajectories ($N=25$ replications) of
      $\AGMMN_{\nkrn}$s trained with fixed number of kernels %
      and median validation loss trajectories ($N=25$ replications) of our
      suggested AGMMN %
      for training data from a Clayton copula with pairwise Kendall's tau
      $\tau = 0.25$ (top left), $\tau = 0.5$ (top right), a Gumbel copula with
      $\tau = 0.25$ (bottom left) and $\tau = 0.50$ (bottom right) in $d = 100$
      dimensions. The vertical dotted lines indicate, on average, a change of
      training phase or termination of training.}
  \label{fig:justifyadaptive_archimedean}
\end{figure}

\begin{figure}[htbp]
  \includegraphics[width=0.49\textwidth]{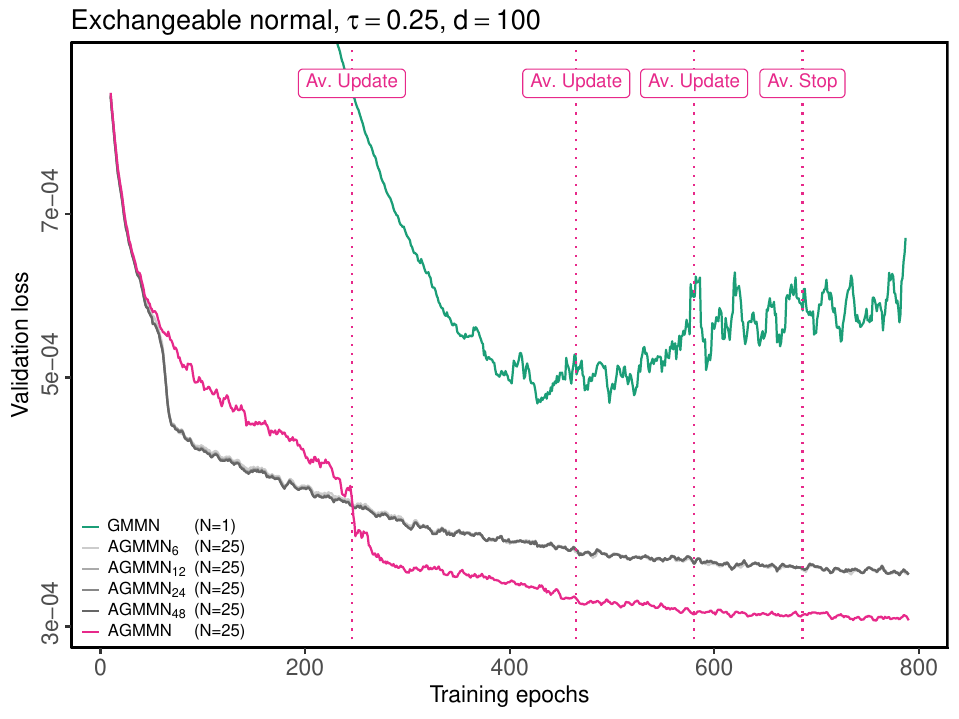}\hfill
  \includegraphics[width=0.49\textwidth]{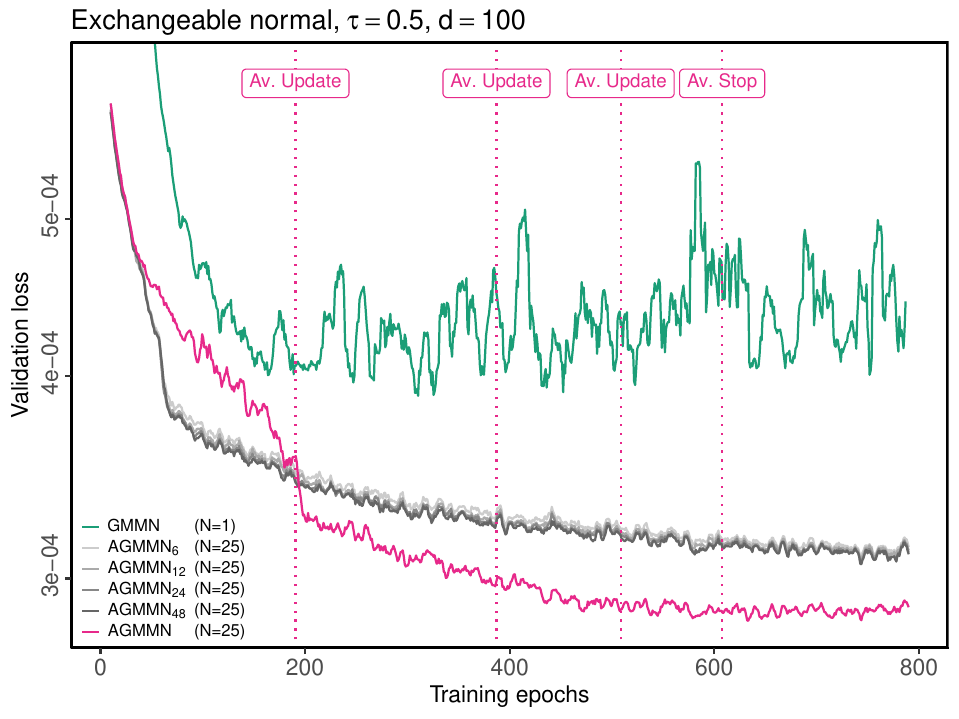}\\[2mm]
  \includegraphics[width=0.49\textwidth]{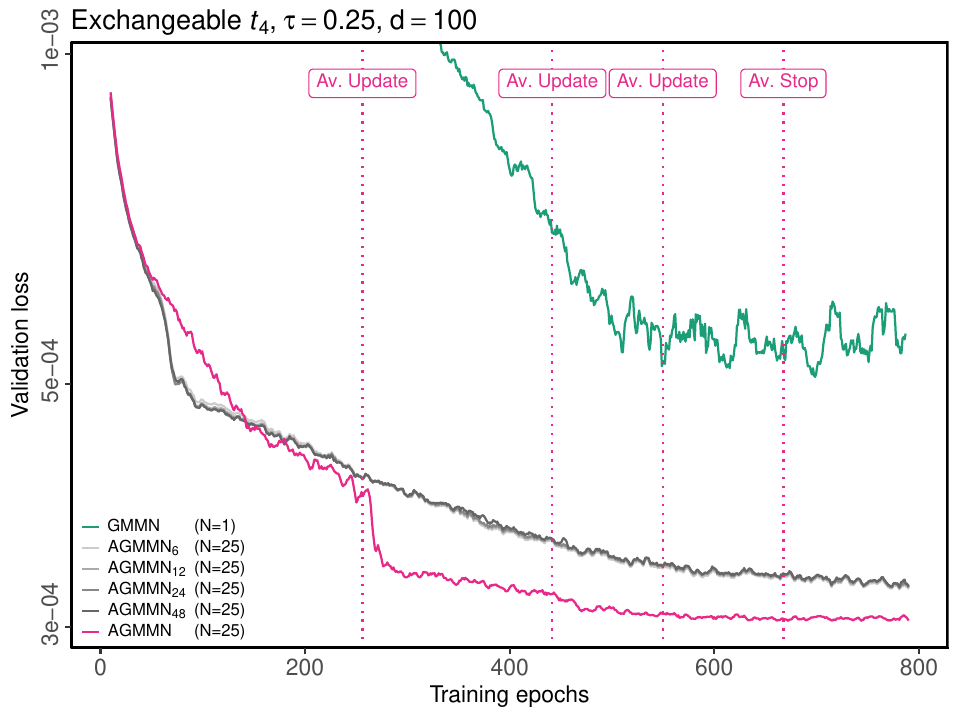}\hfill
  \includegraphics[width=0.49\textwidth]{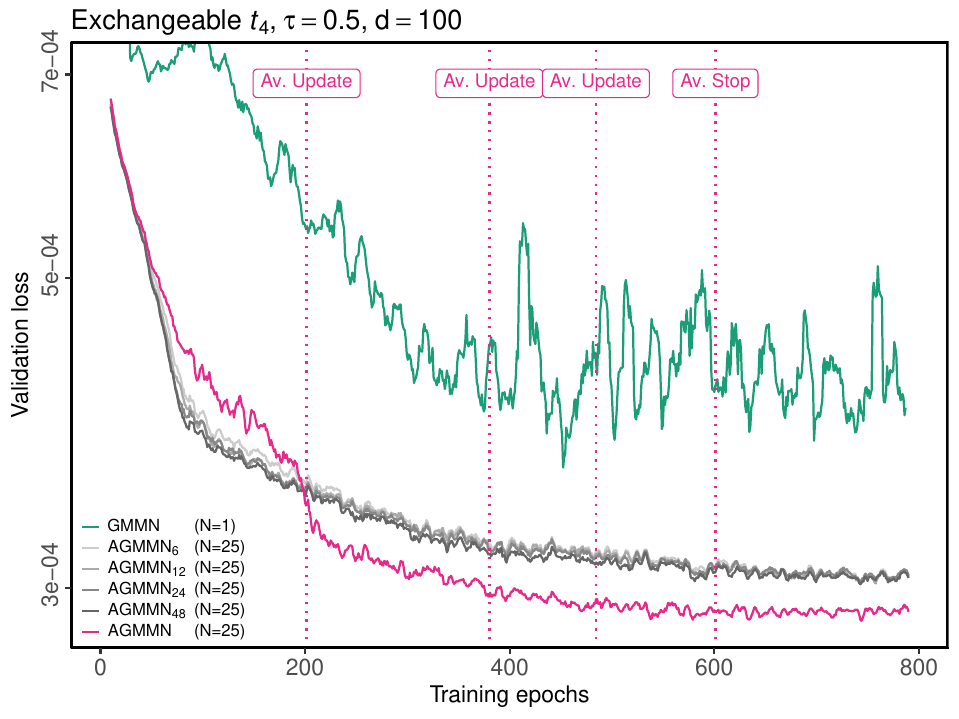}
  \caption{Validation loss trajectory of a GMMN, %
      median validation loss trajectories ($N=25$ replications) of
      $\AGMMN_{\nkrn}$s trained with fixed number of kernels %
      and median validation loss trajectories ($N=25$ replications) of our suggested
      AGMMN %
      for training data from an exchangeable normal copula
      with pairwise Kendall's tau $\tau = 0.25$ (top left), with $\tau = 0.5$
      (top right), an exchangeable $t_4$ copula with $\tau = 0.25$ (bottom left)
      and with $\tau = 0.50$ (bottom right) in $d = 100$ dimensions.
      The vertical dotted lines indicate, on average, a
      change of training phase or termination of training.}
  \label{fig:justifyadaptive_elliptical}
\end{figure}

\begin{figure}[htbp]
  \includegraphics[width=0.49\textwidth]{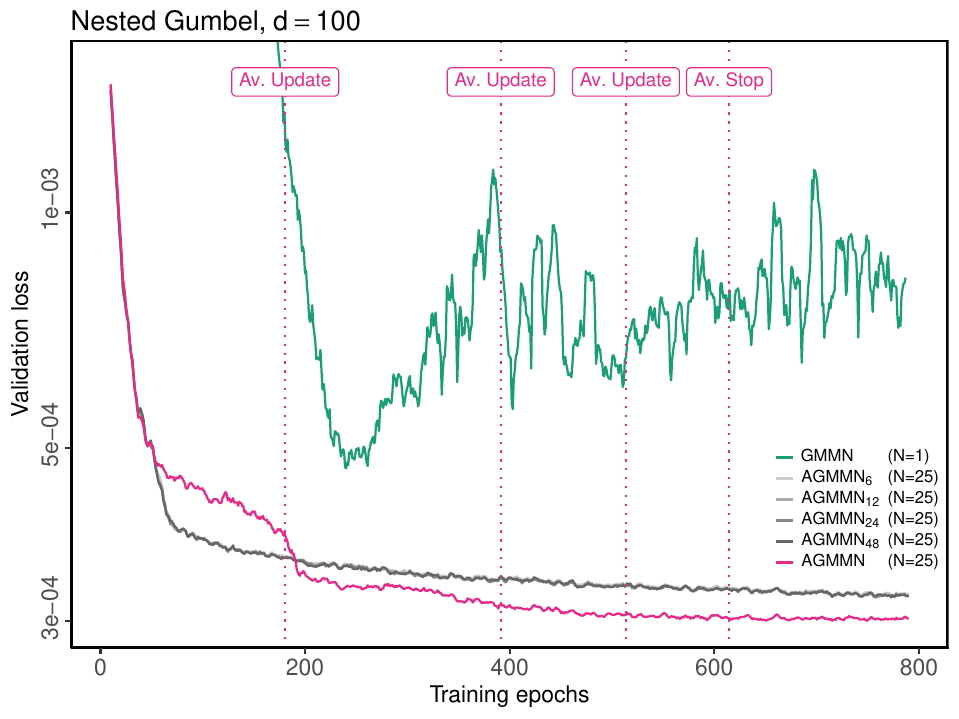}\hfill
  \includegraphics[width=0.49\textwidth]{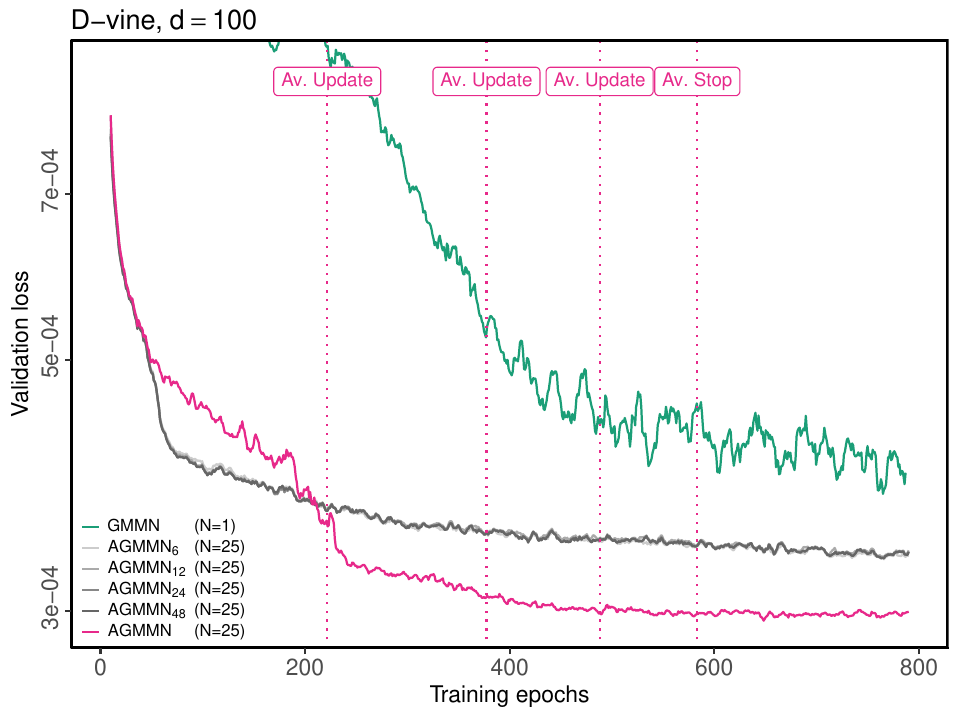}
  \caption{Validation loss trajectory of a GMMN, %
      median validation loss trajectories ($N=25$ replications) of
      $\AGMMN_{\nkrn}$s trained with fixed number of kernels %
      and median validation loss trajectories ($N=25$ replications) of our suggested
      AGMMN %
      for training data from a nested Gumbel copula (left) and a D-Vine copula (right), as specified in
      Section~{\ref{subsec:setup}}, in $d = 100$ dimensions.
      The vertical dotted lines indicate, on average, a
      change of training phase or termination of training.}
  \label{fig:justifyadaptive_flexible}
\end{figure}

To take into account the parameter uncertainty introduced by the stochastic
nature of the training procedure of the neural network's parameters, we repeat
the training of AGMMNs and $\AGMMN_{\nkrn}$s $N = 25$ times, compute the pointwise median of the resulting $N$
validation loss trajectories and plot them for all considered copulas; see
Figures~\ref{fig:justifyadaptive_archimedean},
\ref{fig:justifyadaptive_elliptical} and \ref{fig:justifyadaptive_flexible}.
All figures also include the validation loss trajectory
of a single training of a GMMN for comparison.

There are clear results visible across all considered setups. First, the
performance of the GMMN is far off from all other models (hence also the single
trajectory, there was no point in repeated training of this model), which shows
the advantage of our suggested adaptive training procedure, independently of
whether $\nkrn$ is fixed (as for $\AGMMN_{\nkrn}$s) or chosen adaptively (as for
$\AGMMN$).  Second, the $\AGMMN$ generally converges faster than the
$\AGMMN_{\nkrn}$s.  Third, the $\AGMMN$ generally achieves a lower minimum in
validation loss than the $\AGMMN_{\nkrn}$s.  Fourth, on average, AGMMNs perform
three bandwidth updates before stopping training, which, according to our early
stopping mechanism, means that in each case the first two updates have a
significant positive effect on lowering the validation loss, while the third
update does not; see Algorithm~\ref{alg:AGMMN}
Steps~\ref{alg:stop:updating} and \ref{alg:stop:entirely:or:update:bandwidths}.
In particular, as seen from Figures~\ref{fig:justifyadaptive_archimedean},
\ref{fig:justifyadaptive_elliptical} and \ref{fig:justifyadaptive_flexible}, the
effect of the first update is prominent even in the median validation loss
trajectories, so even after accounting for the parameter uncertainty induced by
the training procedure. Fifth, despite being based on a different number of
kernels $\nkrn$, the $\AGMMN_{\nkrn}$s produce trajectories that are largely
overlapping, that is, on average, it hardly makes any difference to vary $\nkrn$
unless, as we suggest, the bandwidths are adaptively updated during training, as
done for the $\AGMMN$. Sixth, for the same copula family, AGMMNs tend to stop
training later for pairwise Kendall's tau $\tau = 0.25$ than for $\tau =
0.5$, %
which indicates that copulas with weaker strengths of concordance are more
challenging to learn. Lastly, the difference between the smallest validation
loss achieved by GMMNs and AGMMNs tends to be larger for $\tau = 0.25$ than for
$\tau = 0.5$, which indicates that AGMMNs can be especially useful under weaker
dependence and thus more challenging learning situations.

\subsection{Assessment of AGMMN samples}
\begin{figure}[htbp]
  \includegraphics[width=0.48\textwidth]{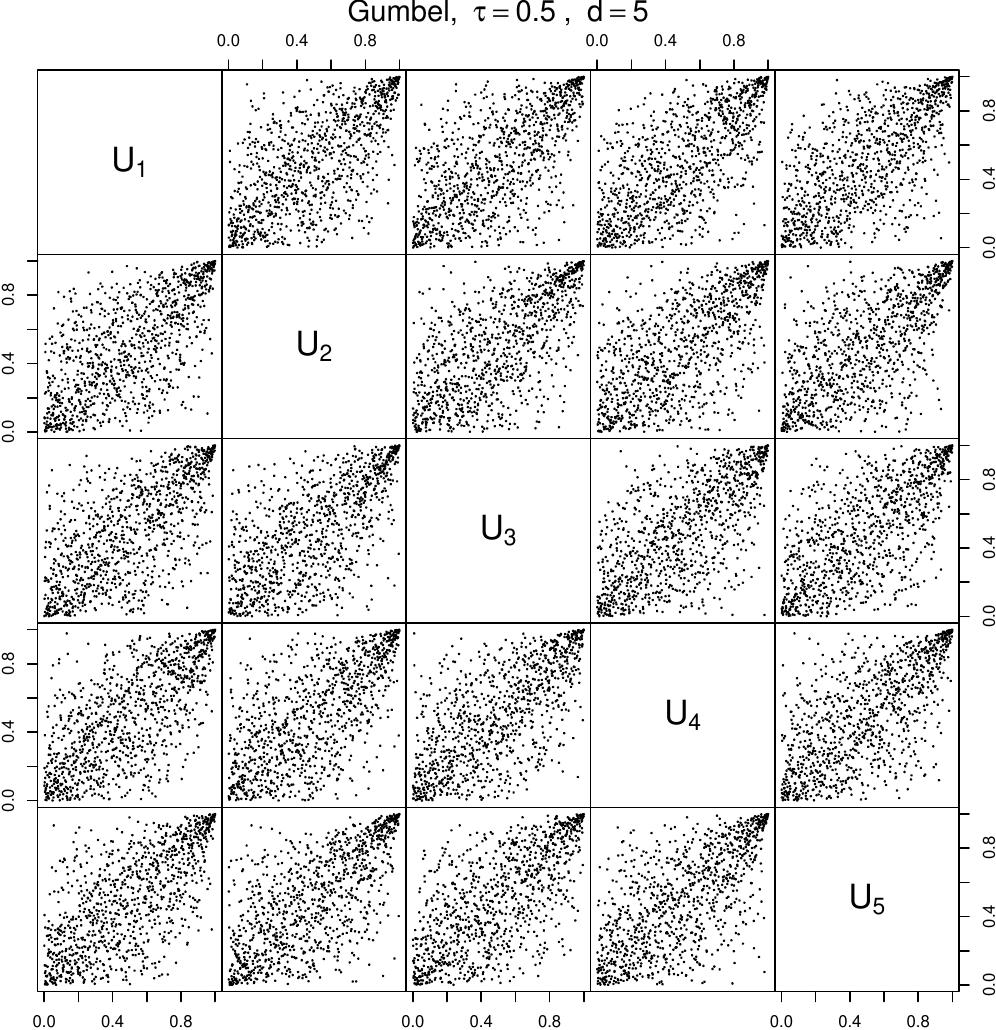}\hfill
  \includegraphics[width=0.48\textwidth]{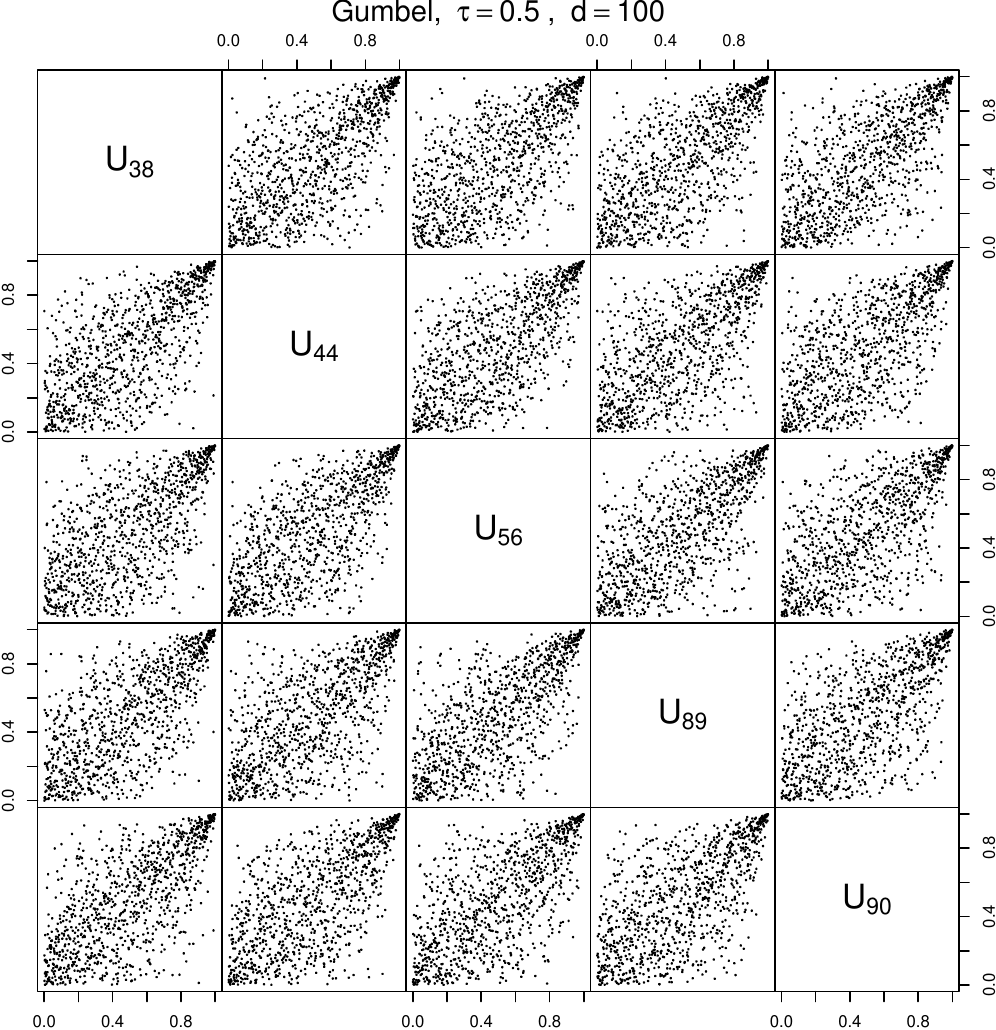}%
  \caption{
    Scatter-plot matrix of a sample of size $\ngen = 1000$ from
    a five-dimensional Gumbel copula with pairwise Kendall's tau $\tau = 0.5$ (left),
    and scatter-plot matrix of five randomly selected component samples from an
    AGMMN trained on a 100-dimensional Gumbel copula sample with the same pairwise Kendall's tau
    (right).
  }
  \label{fig:visuals}
\end{figure}

We now conduct a visual assessment of AGMMN-generated samples.  The left of
Figure~\ref{fig:visuals} shows a scatter-plot matrix of a sample of size $1000$
from a five-dimensional Gumbel copula with pairwise Kendall's tau $\tau=0.5$. As
a comparison, the right of this figure shows, a scatter-plot matrix of five
randomly selected component samples (out of 100) generated by an AGMMN trained
on a 100-dimensional Gumbel copula sample with the same pairwise Kendall's tau.
Due to exchangeability, all pairs of component samples from a fixed Gumbel
copula should be equal in distribution, which can indeed be seen from both plots
of Figure~\ref{fig:visuals}, together with the typical clustering of samples in
the upper-right joint tail induced by upper tail dependence of any Gumbel copula
with positive Kendall's tau. Given the similarity of the two plots, there is,
preliminarily, no indication that the AGMMN-generated sample would not
come from the target Gumbel copula.

\begin{figure}[htbp]
  \centering
  \includegraphics[width=\textwidth]{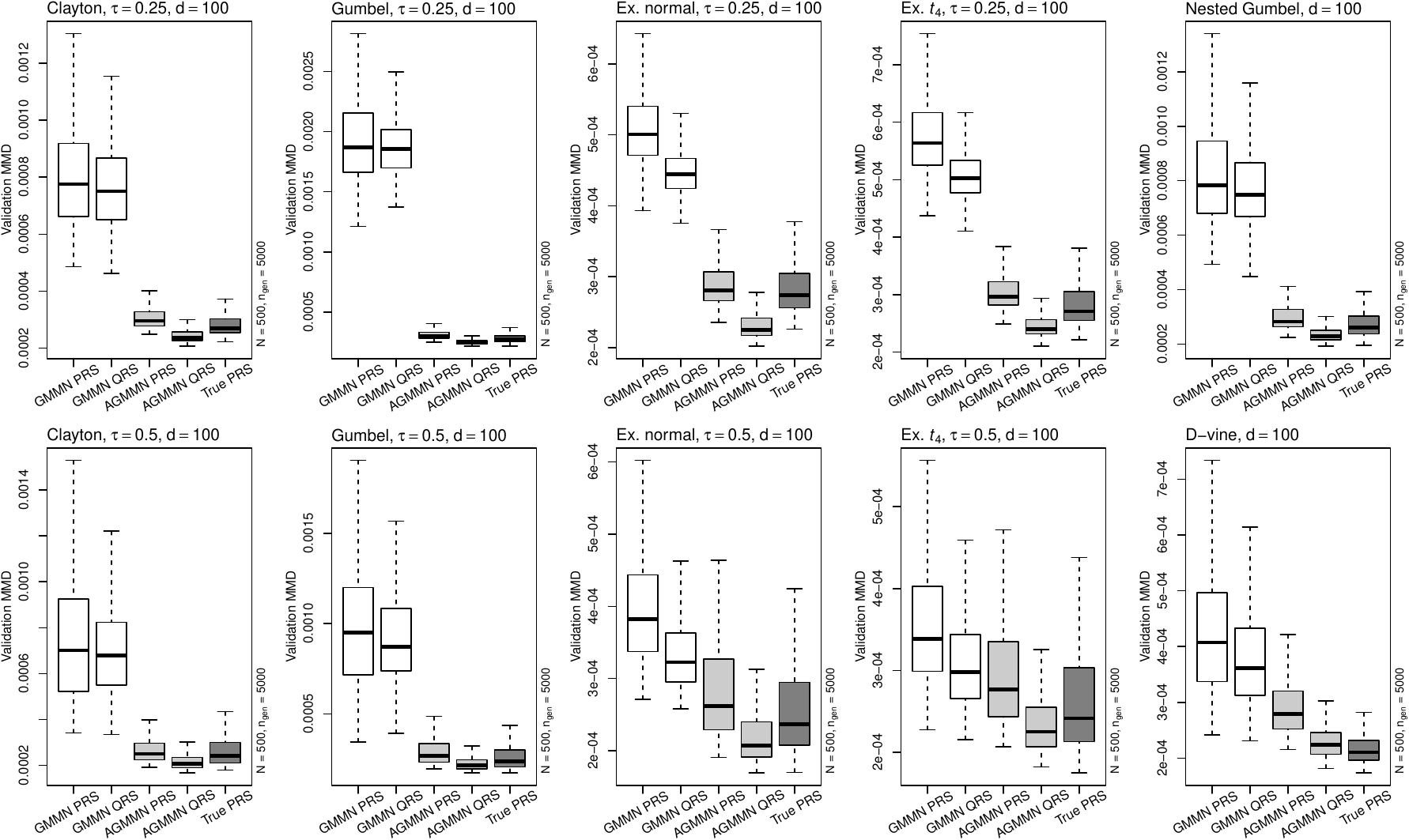}%
  \caption{Boxplots of $N=500$ realizations of the validation MMD~\eqref{eq:MMD}
    based on $\hval$ in~\eqref{eq:hval}, where, for each realization, a new PRS
    of size $\ngen = 5000$ from the true copula is generated and compared with a
    GMMN pseudo-random sample (GMMN PRS), a GMMN quasi-random sample (GMMN QRS),
    an AGMMN pseudo-random sample (AGMMN PRS), an AGMMN quasi-random sample
    (AGMMN QRS) and another pseudo-random sample from the true copula (TRUE
    PRS), and this for Clayton copulas (first column), Gumbel copulas (second
    column), exchangeable normal copulas (third column), exchangeable $t_4$
    copulas (fourth column) with pairwise Kendall's tau $\tau = 0.25$ (top row,
    first four columns) and $\tau = 0.5$ (bottom row, first four columns), and
    for a nested Gumbel copula (fifth column, top) and a D-vine copula (fifth
    column, bottom) as specified in Section~{\ref{subsec:setup}}.}
  \label{fig:mmd_boxplots}
\end{figure}

Next, we assess the quality of AGMMN-generated samples using boxplots of $N=500$
realizations of validation MMD values~\eqref{eq:MMD} with bandwidths as
in~\eqref{eq:hval}; note that $\hval$ differs from the vector of bandwidths used
for computing the MMD training loss for all considered
models. Figure~\ref{fig:mmd_boxplots} shows the boxplots, where, for each
validation MMD realization, we first generate a pseudo-random sample (PRS) of
size $\ngen = 5000$ from the true copula and then compute its validation MMD
against a GMMN PRS, a GMMN quasi-random sample (QRS), an AGMMN PRS, an AGMMN QRS
and another PRS from the true copula (indicated by ``TRUE PRS'').

We see that for all considered copulas, the AGMMN PRS can produce validation MMD
values similar to those of the true copula PRS, but neither the GMMN PRS nor the GMMN
QRS can. Furthermore, the AGMMN QRS can produce validation MMD values lower than
those of the true copula PRS. Moreover, the magnitude by which the AGMMNs
outperform the GMMNs is larger when the dependence is weaker. This again
implies that AGMMNs are preferable to GMMNs in more challenging learning situations.

\section{Analysis of the AGMMN RQMC estimator}\label{sec:rqmc}
We now analyze the variance-reduction properties of three AGMMN RQMC estimators
of quantities $\mu$ of interest that can be written in the form
\begin{align}
  \muAGMMNRQMC = \frac{1}{\ngen} \sum_{i=1}^{\ngen} \Psi_{l,i}\bigl(f_{\hat{\bm{\theta}}}(F_{\bm{Z}}^{-1}(\tilde{\bm{v}}_1)),\dots,f_{\hat{\bm{\theta}}}(F_{\bm{Z}}^{-1}(\tilde{\bm{v}}_{\ngen}))\bigr),\quad l=1,2,3,\label{eq:RQMC:estimator:form}
\end{align}
where $\{\tilde{\bm{v}}_i\}_{i=1}^{\ngen}$ is a digital-shifted $d$-dimensional
Sobol' sequence, $f_{\hat{\bm{\theta}}}$ is a trained AGMMN (whose outputs are
transformed to pseudo-observations as in Algorithm~\ref{alg:QRNG:AGMMN}) and the
functionals $\Psi_{l,i}$ (specified later) in~\eqref{eq:RQMC:estimator:form}
share the same mean for different $i=1,\dots,\ngen$. %
We compare the AGMMN RQMC estimator $\muAGMMNRQMC$ with the
GMMN RQMC estimator $\muGMMNRQMC$, the MC estimator $\mucopMC$ based on copula
pseudo-random samples, and the RQMC estimator
$\mucopRQMC$ %
based on copula quasi-random samples. For each model, each type of estimator and
each of the three functionals considered, \emph{sample standard deviation (SSD)} estimates
$\hat{s}_{\ngen}$ are computed for all sample sizes
$\ngen \in \{2^{10},2^{10.5} \dots, 2^{20}\}$ based on $N = 25$ RQMC point sets
$\tilde{P}_{\ngen}=\{\tilde{\bm{v}}_1,\dots,\tilde{\bm{v}}_{\ngen}\}$. Then
coefficients $\alpha$ are computed by fitting the linear regression
$\hat{s}_{\ngen}=\alpha n_{\ngen}+\eps_{\ngen}$,
$\ngen \in \{2^{10},2^{10.5} \dots, 2^{20}\}$.  The regression coefficients
$\alpha$ serve as a numerical measure of the convergence rates of MC and RQMC
estimators.

The first of the three quantities we consider to estimate is the \emph{expected
  shortfall}
\begin{align}
  \mu= \ES_{0.99}(S) = \frac{1}{1 - 0.99} \int_{0.99}^1 F_S^{-1}(u)\,\rd u = \E(S\,|\,S > F_S^{-1}(0.99))\label{eq:es99}
\end{align}
at level $0.99$ of the aggregated loss $S = \sum_{j=1}^d X_j$, where
$\bm{X} = (X_1,\dots,X_d)$ models risk-factor changes with $\N(0,1)$ margins and
different copulas to be detailed later, and where
$F_{S}^{-1}(u) = \inf\{x \in \IR: F_S(x)\ge u\}$ is the quantile function of the
distribution function $F_S$ of $S$. Expected shortfall is an important risk measure and
is used to determine an amount of risk capital to put aside now in order to
account for future losses in the realm of quantitative risk management. Omitting
additional notation for the fact that we build pseudo-observations in
Algorithm~\ref{alg:QRNG:AGMMN} Step~\ref{alg:QRNG:AGMMN:postprocess}, this
corresponds to the case $l=1$ in~\eqref{eq:RQMC:estimator:form}, for
$\Psi_{1,i}(\bm{u}_1,\dots,\bm{u}_{\ngen})=\frac{\ngen}{\sum_{k=1}^{\ngen}\I_{\{s_k>s_{(\lceil
	0.99\ngen\rceil)}\}}}s_i\I_{\{s_i>s_{(\lceil 0.99\ngen\rceil)}\}}$ with $s_i=\sum_{j=1}^d\Phi^{-1}(u_{i,j})$,
  $i=1,\dots,\ngen$, and $s_{(1)}\le\dots\le s_{(\ngen)}$ denoting the order statistics as usual.

The second quantity we consider to estimate is the \emph{expected shortfall
  contribution}
\begin{align*}
  \mu = \AC_{1, 0.99} = \E(X_1\,|\, S > F_S^{-1}(0.99))
\end{align*}
of the (without loss of generality) first component $X_1$ of $\bm{X}$ in the
context of capital allocation in quantitative risk management. Expected
shortfall contributions are used to determine how to allocate risk capital, for
example a computed expected shortfall value, among several business lines.  Also
here, $\bm{X}$ is assumed to have $\N(0,1)$ margins and, as mentioned before,
the copulas of $\bm{X}$ considered are specified later.
The functional corresponding to this setup is
$\Psi_{2,i}(\bm{u}_1,\dots,\bm{u}_{\ngen})=\frac{\ngen}{\sum_{k=1}^{\ngen}\I_{\{s_k>s_{(\lceil
      0.99\ngen\rceil)}\}}}\Phi^{-1}(u_i)\I_{\{s_i>s_{(\lceil 0.99\ngen\rceil)}\}}$, $i=1,\dots,\ngen$.

As third quantity, we consider an application from finance, namely to estimate the
expected payoff
\begin{align}
  \mu = \E\biggl(\exp(-r(T - t))\max\biggl\{\,\biggl(\frac{1}{d}\sum_{j=1}^d S_{T, j}\biggr)-K,0\biggr\}\biggr)\label{eq:bc}
\end{align}
of a European basket call option, where $r$ denotes the continuously compounded
annual risk-free interest rate (chosen as 0.01), $t$ denotes the current time
point (chosen as 0), $T$ the maturity in years (chosen as 1) and $K$ the strike
price of the option (chosen as 1.01). We assume a Black--Scholes framework for
the marginal stock prices $(S_{T, 1}, \dots, S_{T,d})$ at maturity $T$, so
assume that $S_{T, j}$ follows a log-normal distribution
$\LN(\log(S_{t,j}) + (r - \sigma_j^2/2)(T - t), \sigma_j^2(T - t))$ with distribution function $F_{\LN,j}$, where
$S_{t,j}$ denotes the last available stock price of the $j$th constituent in the basket and
$\sigma_j$ denotes its volatility. We take
$(S_{t,1},\dots,S_{t,d}) = (1,\dots,1)$ and $(\sigma_1, \dots, \sigma_d)$ to be
an equidistant sequence between $0.01$ and $0.025$. The dependence structure of
$(S_{T, 1}, \dots, S_{T,d})$ is modeled with various copulas to be detailed
later. The corresponding functional is $\Psi_{3,i}(\bm{u}_1,\dots,\bm{u}_{\ngen})=\exp(-r(T - t))\max\{(\frac{1}{d}\sum_{j=1}^dF_{\LN,j}^{-1}(u_{i,j}))-K,0\}$, which, in this case, only depends on its $i$th component $\bm{u}_i$ %
and so the AGMMN RQMC estimator in~\eqref{eq:RQMC:estimator:form} is a classical
RQMC estimator. %

The variance-reduction capabilities of the GMMN RQMC estimator $\muGMMNRQMC$ and
the AGMMN RQMC estimator $\muAGMMNRQMC$ depend on the low-discrepancy properties
of the underlying QMC point set. The quality of the Sobol' sequence in such high
dimensions we consider in this work may thus play a role in the deterioration of
the variance-reduction capabilities of $\muGMMNRQMC$ and $\muAGMMNRQMC$; see
\cite{hofertprasadzhu2021a} where this phenomenon was observed in dimensions as
low as $d = 10$, but no reason was given. In Appendix~A.1, we thus
simulate the fluctuation of points of a digital-shifted Sobol' sequence in
a hyperrectangle as the dimension $d$ increases. We demonstrate that an increase
in variance is expected in higher dimensions even though the expected number of
points in the hyperrectangle remains the same across different
dimensions. Having observed this behavior of randomized Sobol' sequences in higher
dimensions, we also experiment with prior dimensions $\dpri$ lower than the data
dimension $d$, while keeping the training setup as specified in
Section~\ref{subsec:setup}.

\begin{figure}[htbp]
	\includegraphics[width=0.3\textwidth]{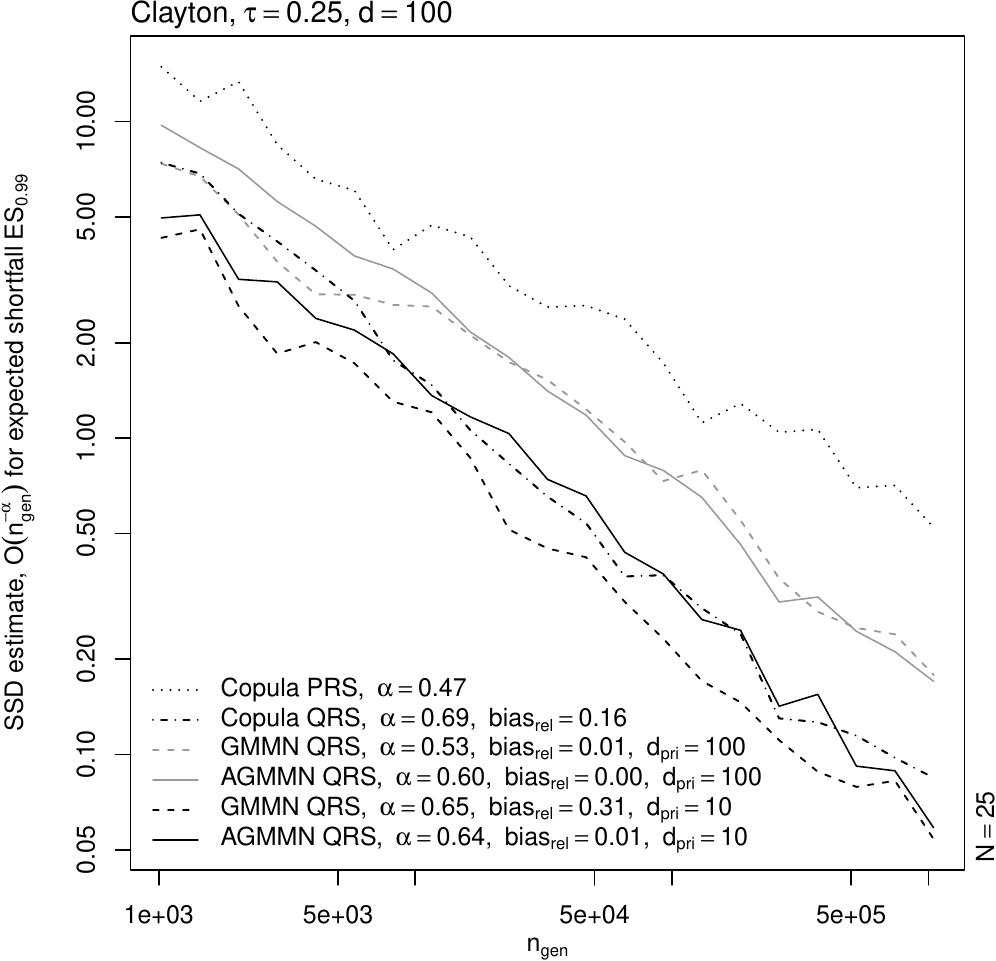}\hfill
	\includegraphics[width=0.3\textwidth]{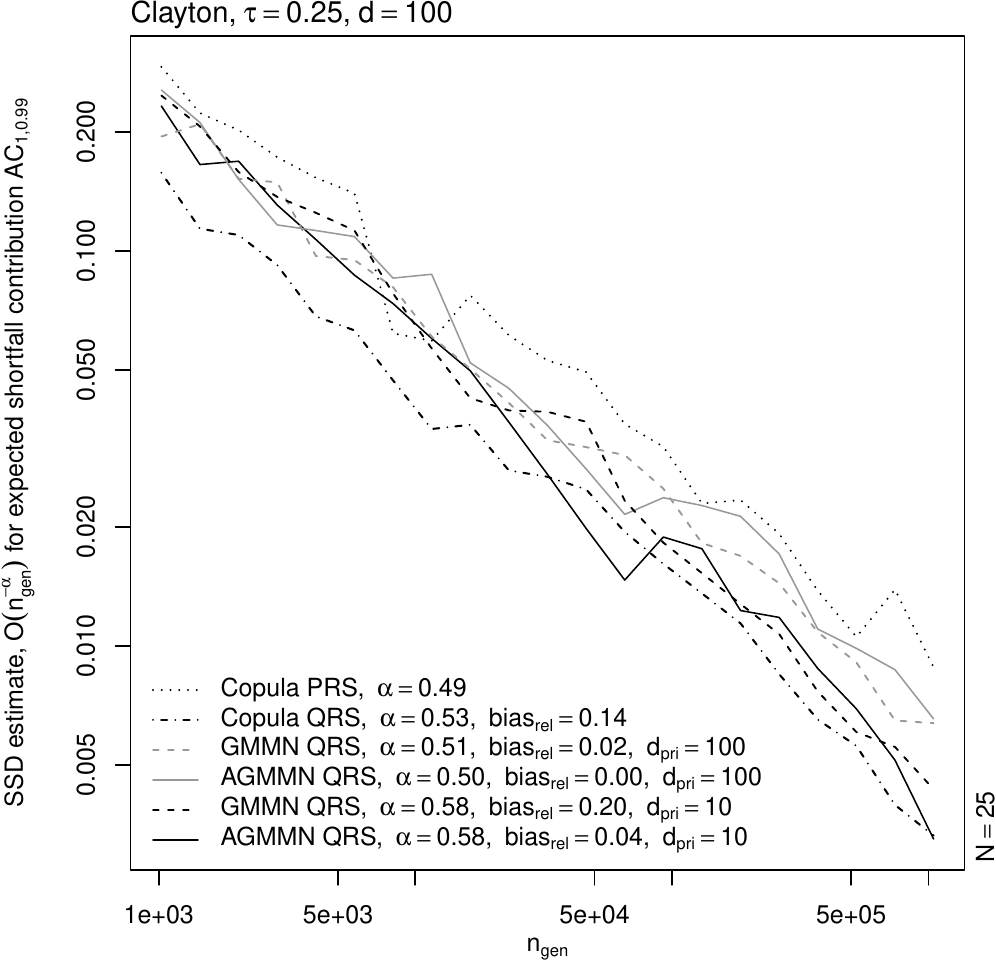}\hfill
	\includegraphics[width=0.3\textwidth]{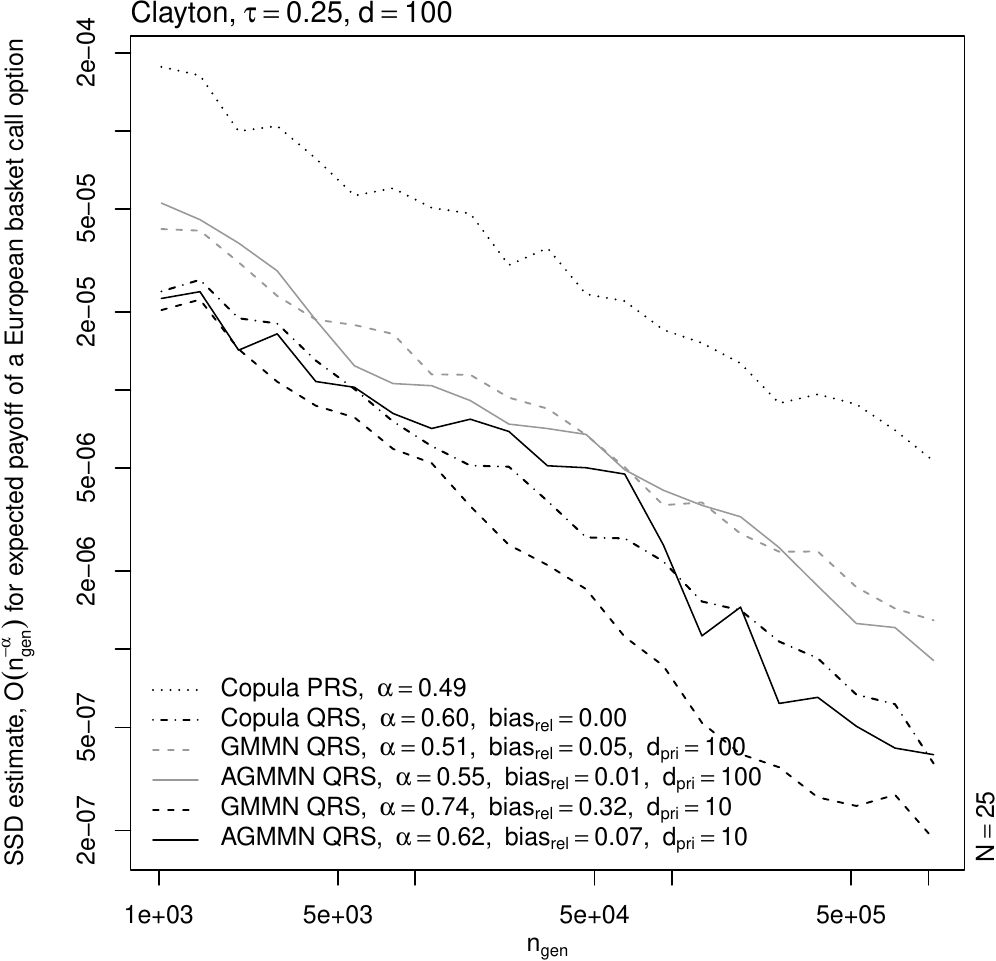}\\[1.5mm]
	\includegraphics[width=0.3\textwidth]{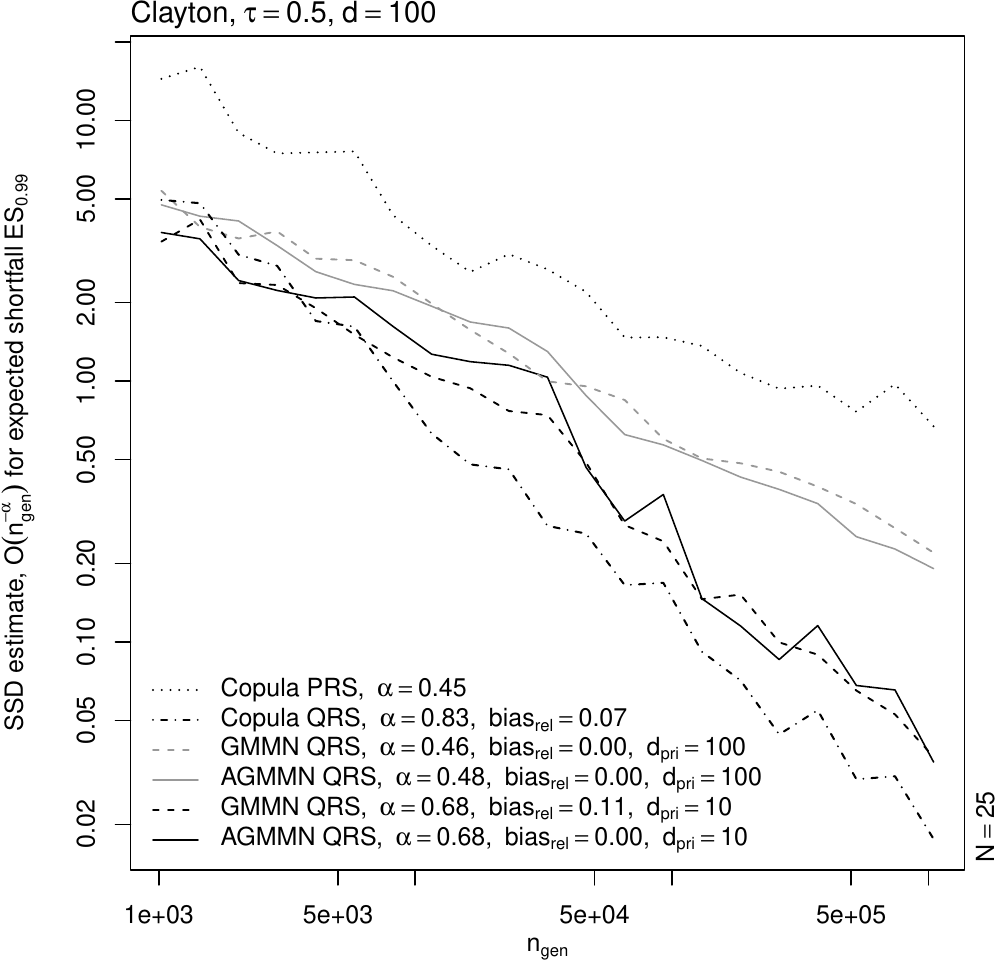}\hfill
	\includegraphics[width=0.3\textwidth]{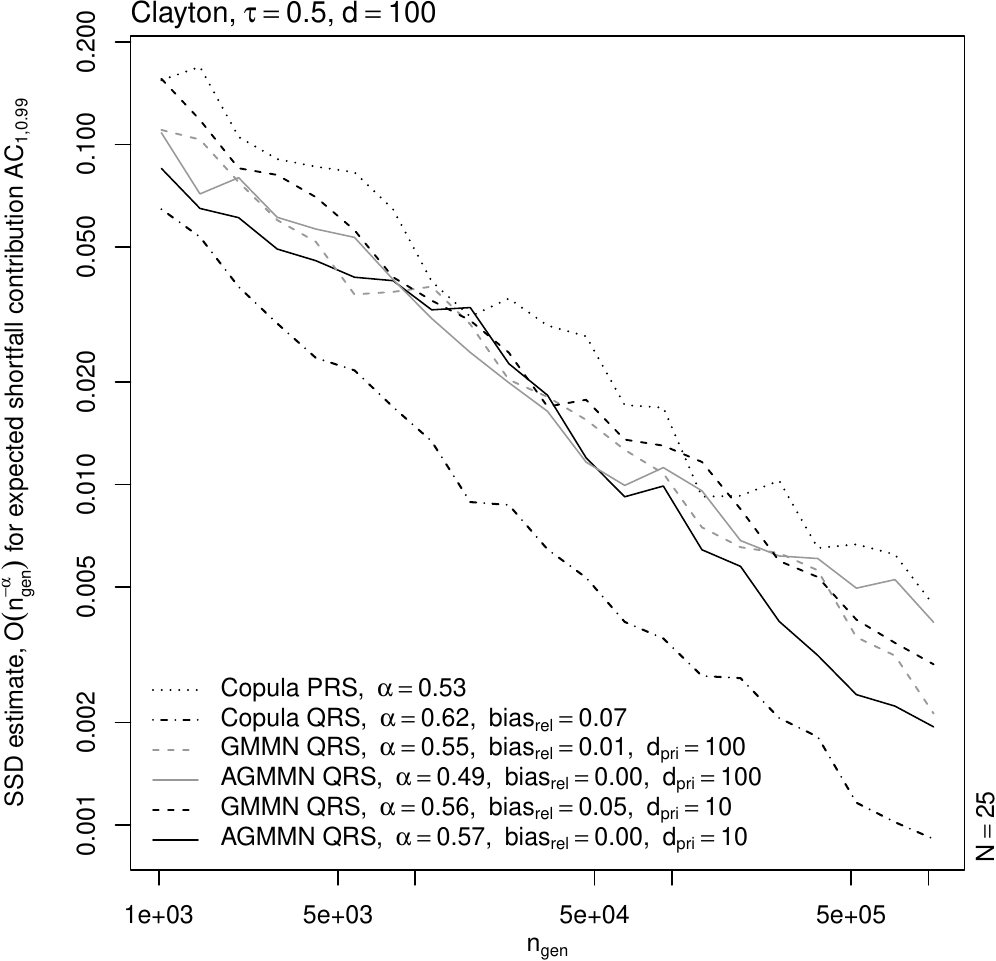}\hfill
	\includegraphics[width=0.3\textwidth]{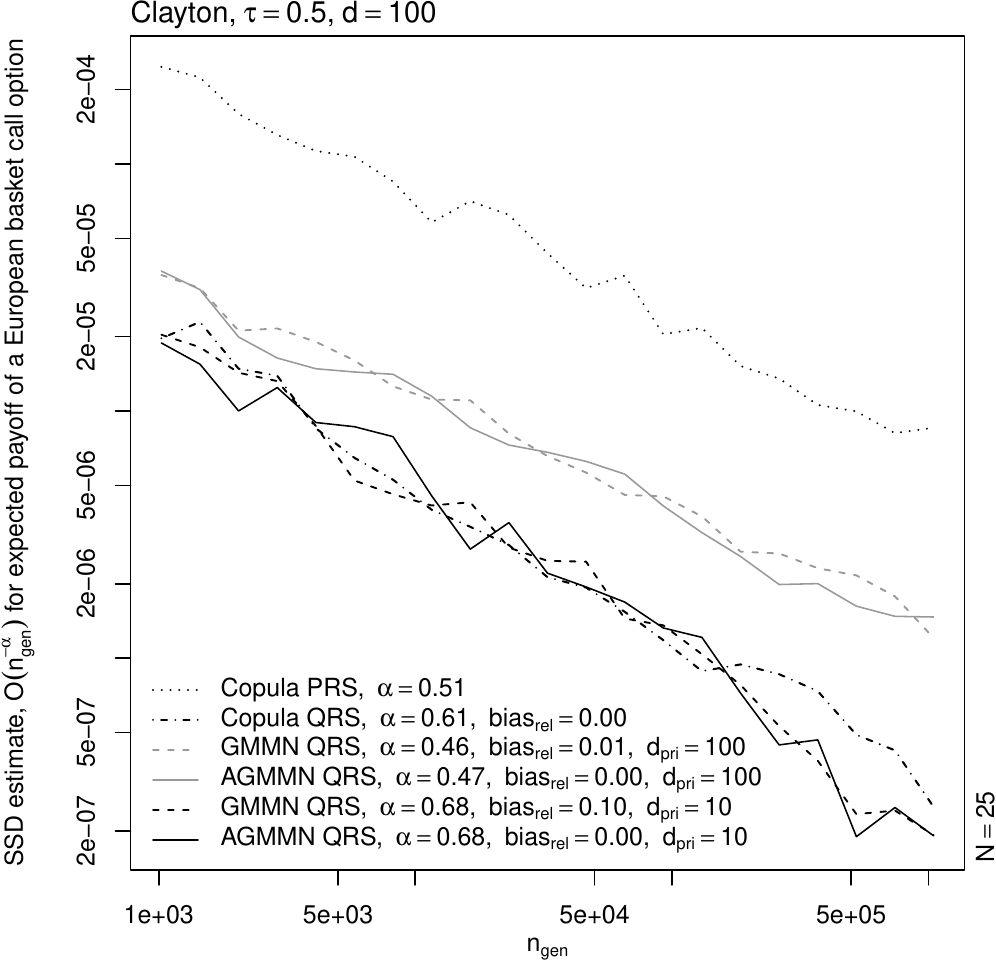}\\[1.5mm]
	\includegraphics[width=0.3\textwidth]{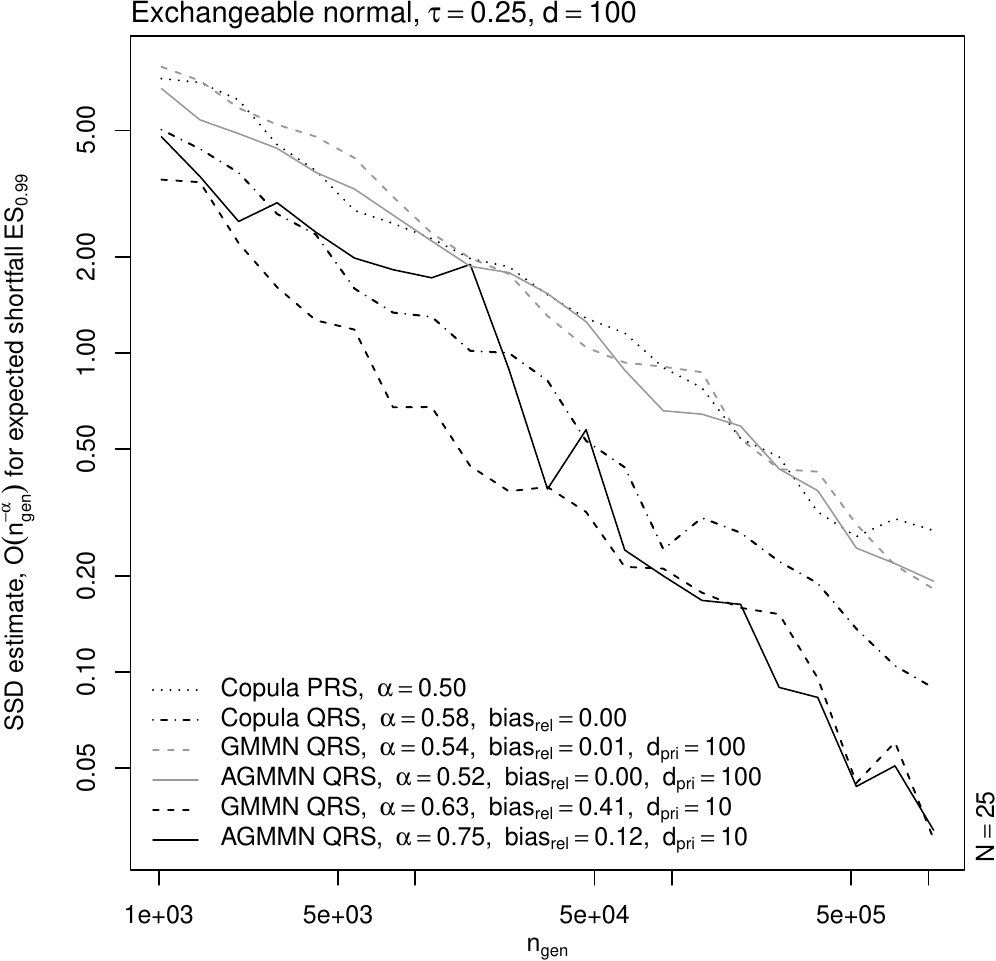}\hfill
	\includegraphics[width=0.3\textwidth]{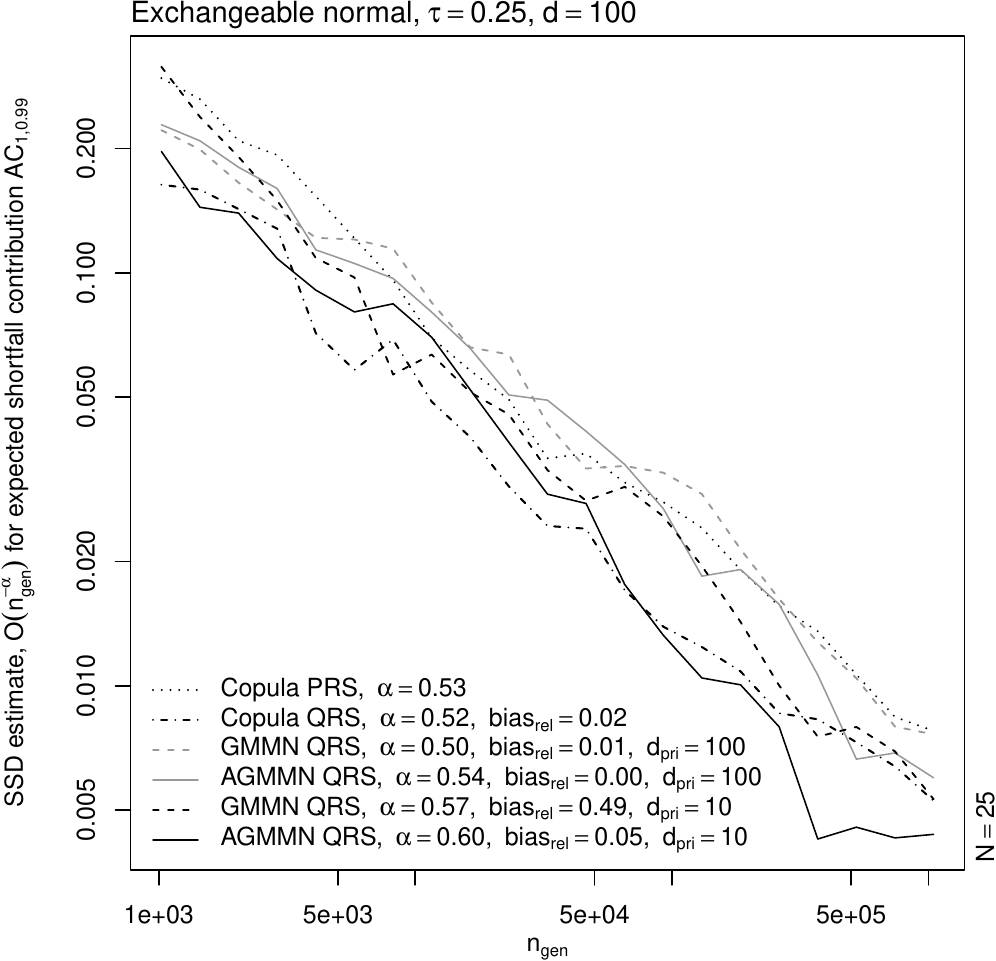}\hfill
	\includegraphics[width=0.3\textwidth]{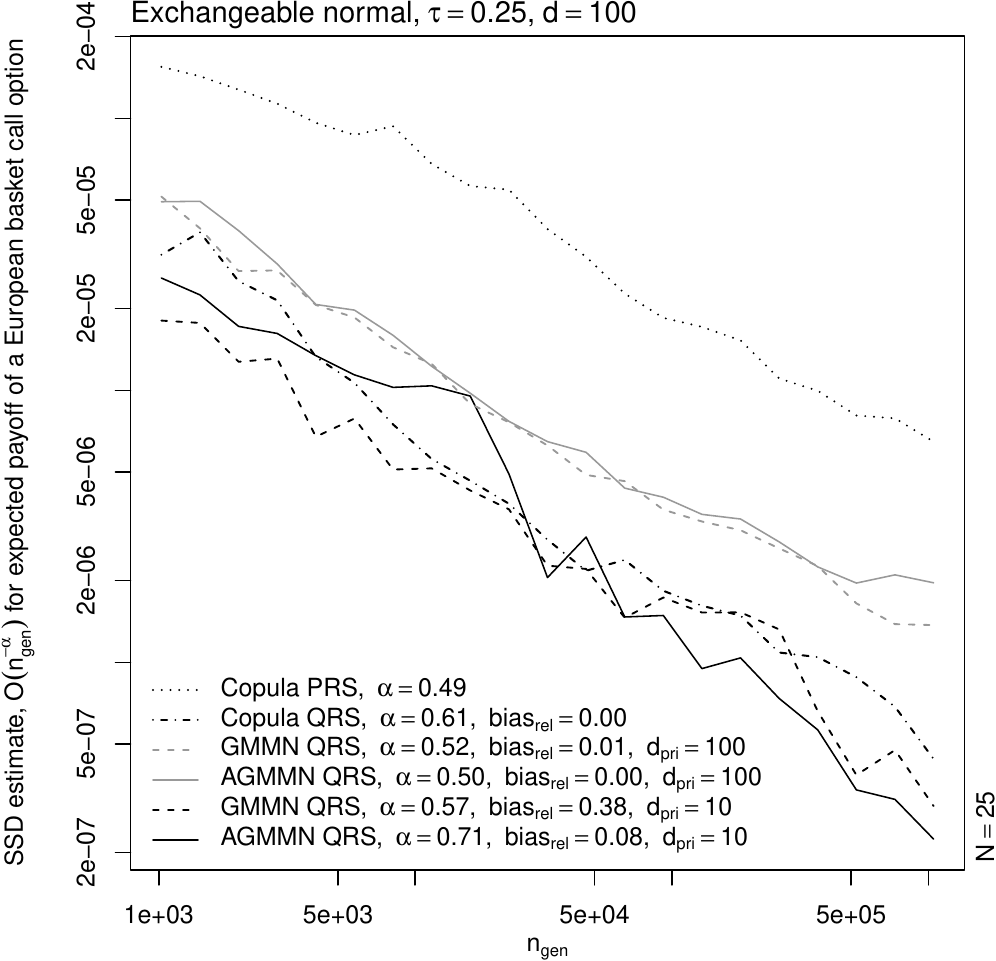}\\[1.5mm]
	\includegraphics[width=0.3\textwidth]{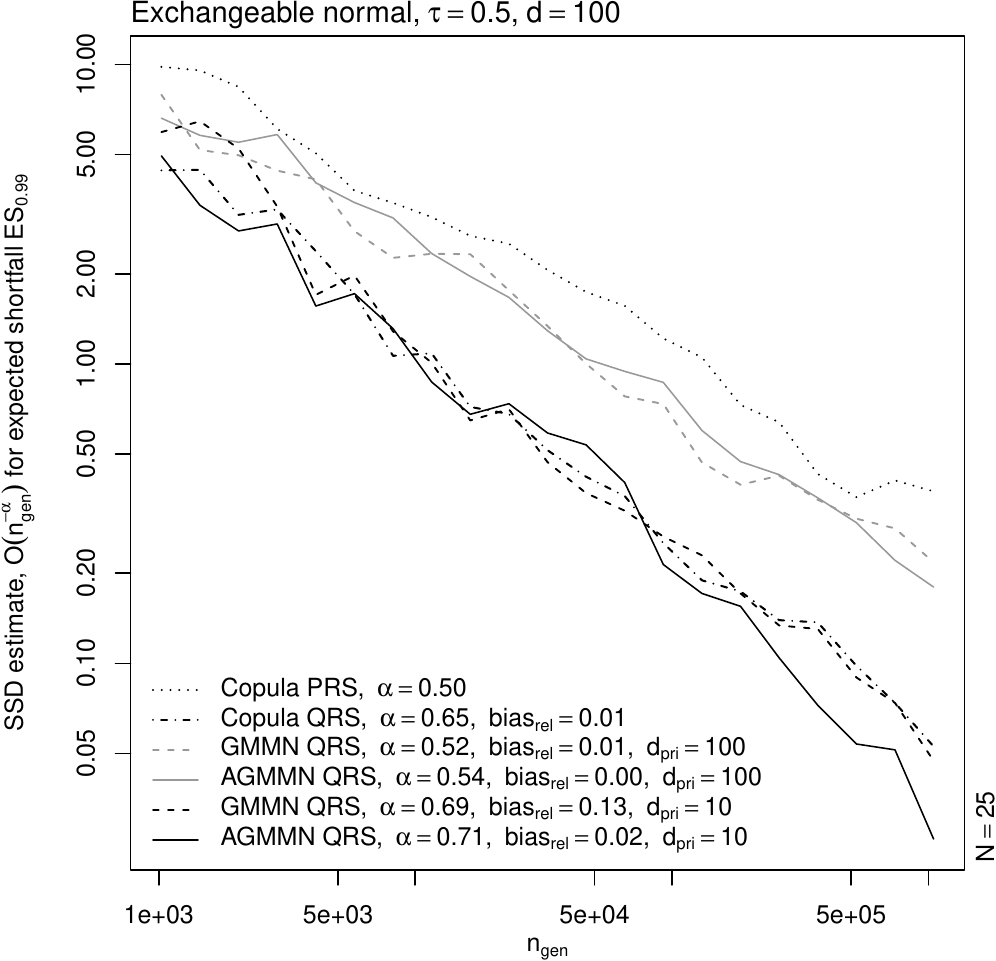}\hfill
	\includegraphics[width=0.3\textwidth]{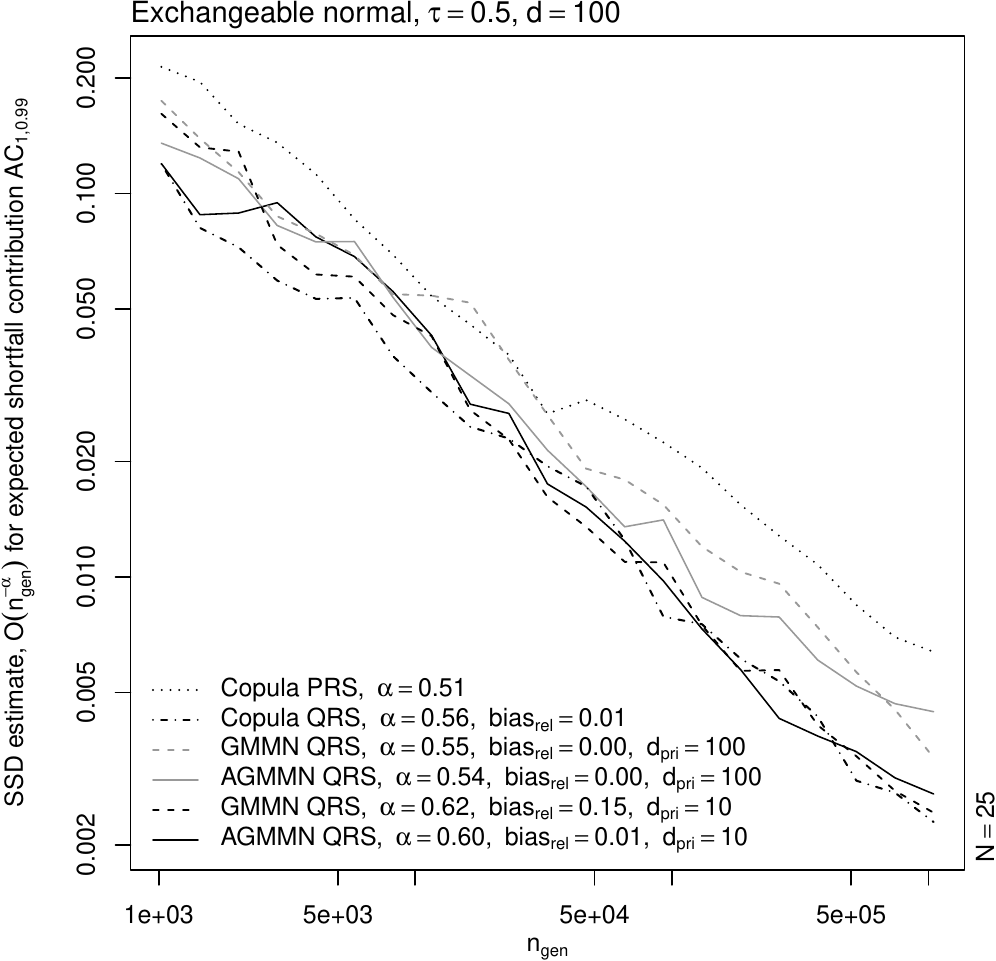}\hfill
	\includegraphics[width=0.3\textwidth]{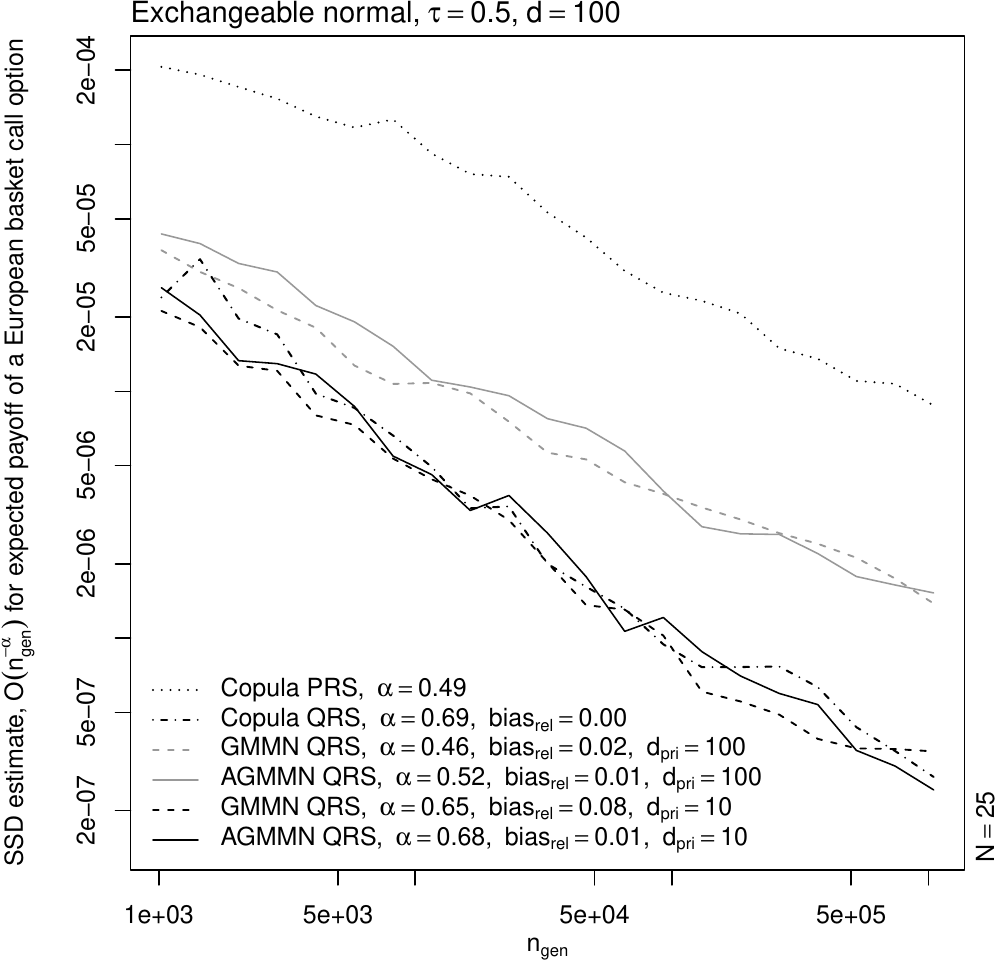}
	\caption{SSDs when estimating $\mu$ (left column: $l=1$; center column:
          $l=2$; $l=3$: right column) via the copula MC estimator $\mucopMC$,
          the copula RQMC estimator $\mucopRQMC$ and the neural network RQMC
          estimators $\muGMMNRQMC$ and $\muAGMMNRQMC$ with prior dimensions
          $\dpri = 10$ and $\dpri = 100$ for Clayton copulas with pairwise
          Kendall's tau $\tau = 0.25$ (row 1) and $\tau = 0.5$ (row 2) and
          exchangeable normal copulas with pairwise Kendall's tau $\tau = 0.25$
          (row 3) and $\tau = 0.5$ (row 4).}
	\label{fig:RQMC}
\end{figure}

Figure~\ref{fig:RQMC} shows plots of SSDs when estimating the
three quantities $\mu$ (left column: $l=1$; center column: $l=2$; right column:
$l=3$) according to the aforementioned setup and for 100-dimensional Clayton
copulas with $\tau\in\{0.25, 0.5\}$ (first two rows, respectively), as well as for
100-dimensional exchangeable normal copulas with $\tau\in\{0.25, 0.5\}$ (last two
rows, respectively).
We choose Clayton and normal copulas because they are two of the few copula
families that have computationally tractable inverse transformations of
\cite{rosenblatt1952}. They thus allow for natural quasi-random sampling procedures
so that a copula RQMC estimator $\mucopRQMC$ is available to compare against. For
most other copula families, quasi-random sampling is either too time-consuming or
not available (\cite{cambouhofertlemieux2017}, \cite{hofertprasadzhu2021a}).
We also report in the legends of the plots of Figure~\ref{fig:RQMC} the
\emph{absolute relative bias}
\begin{align*}
  \bias_{\text{rel}}(\ngen) = \biggl|\frac{\mudot - \muMC}{\muMC}\biggr|,
\end{align*}
so the relative error at $\ngen = 2^{20}$ in comparison to the copula MC
estimator $\muMC$, where
$\mudot \in \{\mucopRQMC, \muGMMNRQMC, \muAGMMNRQMC\}$.

In the plots, we clearly see a bias-variance trade-off of GMMNs and AGMMNs for
different prior dimensions $\dpri$. For $\dpri = 100$, $\muGMMNRQMC$ and
$\muAGMMNRQMC$ have low bias, with the bias of $\muGMMNRQMC$ being slightly larger,
and the variance estimates of $\muGMMNRQMC$ and $\muAGMMNRQMC$ being of an order
between those of $\mucopMC$ and $\mucopRQMC$. For $\dpri = 10$, variance estimates of
$\muGMMNRQMC$ and $\muAGMMNRQMC$ significantly improve and become comparable with
those of $\mucopRQMC$ due to the lower model complexity and better quality of the
Sobol' sequence in lower dimensions (Appendix~A.1), at the cost of an increased
bias. Specifically, the absolute relative bias of the GMMN estimator $\muGMMNRQMC$
can reach up to 49\%, whereas that of the AGMMN estimator $\muAGMMNRQMC$ is at most
12\%, with the majority of cases being around 5\%. Compared to GMMNs, the improved
learning of AGMMNs can thus keep the bias of $\muAGMMNRQMC$ under control even with
a prior distribution that has a much smaller dimension $\dpri$ than the data
dimension $d$. Also, note that we only have access to $\mucopRQMC$ in very few
cases, hence the general-purpose AGMMN estimator $\muAGMMNRQMC$ applied with an
appropriate prior dimension $\dpri$ achieves a good balance between flexibility,
bias and variance.

\section{A synthetic data example: Dependence implied by S\&P~500 constituents}\label{sec:simulation}
In this section, we show that the improved fit achieved by AGMMNs over GMMNs can
indeed translate to a better performance of MC estimators.

To this end, we consider 7815 daily adjusted closing prices of 50 constituents
of the S\&P~500 from 1985-01-01 to 2015-12-31. The data are available in the \R\
package \texttt{qrmdata} (\cite{qrmdata}); see Appendix~A.2 for details. To
account for temporal dependence, we model each of the 50 marginal time series of
log-returns using an $\ARMA(1,1)$--$\GARCH(1,1)$ model with standardized $t$
innovation distribution and extract the marginal standardized residuals
(\emph{deGARCHing}). We then compute the pseudo-observations of the standardized
residuals which serve as a sample for modeling cross-sectional dependence. These
pseudo-observations are modeled in Section~\ref{sec:prediction}.  In
Section~\ref{sec:simulation} here, however, we fit a $t$ copula to these
pseudo-observations and treat the fitted $t$ copula as the true underlying
model. This way, we get to evaluate how close an estimate is to this
data-implied ground truth. %

With a sample of size $\ntrn = 7815$ from the true fitted $t$ copula model (the
same sample size as the original data), we train GMMNs and AGMMNs in
mini-batches of size $\nbat = 2000$ and compare their MC ($\muGMMNMC$,
$\muAGMMNMC$) and RQMC ($\muGMMNRQMC$, $\muAGMMNRQMC$) estimators of two of the
three quantities $\mu$ considered in Section~\ref{sec:rqmc}, namely the expected
shortfall as in~\eqref{eq:es99} ($l=1$) and the expected payoff of a basket call
option as in~\eqref{eq:bc} ($l=3$). For expected shortfall, the margins of
$\bm{X} = (X_1, \dots, X_d)$ are now the fitted standardized $t$ distributions
obtained from the aforementioned deGARCHing. Trained GMMNs and AGMMNs are then
used as models of the dependence structure of $\bm{X}$.  For the expected payoff
of the basket call option, and in alignment with the real data, we take
$S_{t,j}$ for $t=0$ to be the adjusted closing price of the $j$th constituent on
2015-12-31, and estimate $\sigma_j$ by the SSD $\hat{s}_j$ of the log-returns of
the $j$th constituent over the time period 2014-01-01 to 2015-12-31. We thus
have the margins
$S_{T,j}\sim\LN(\log(S_{t,j}) + (r - \hat{s}_j^2/2)(T - t), \hat{s}_j^2(T - t))$
and the dependence of $(S_{T,1},\dots,S_{T,d})$ is modeled by GMMNs and AGMMNs.
Furthermore, we choose $T=1$ and $r=0.01$ as in Section~\ref{sec:rqmc}, and the
strike price $K$ is chosen 0.5\% higher than the average of all stock prices at
$t=0$.

\begin{figure}[htbp]
  \centering
  \includegraphics[width=0.35\textwidth]{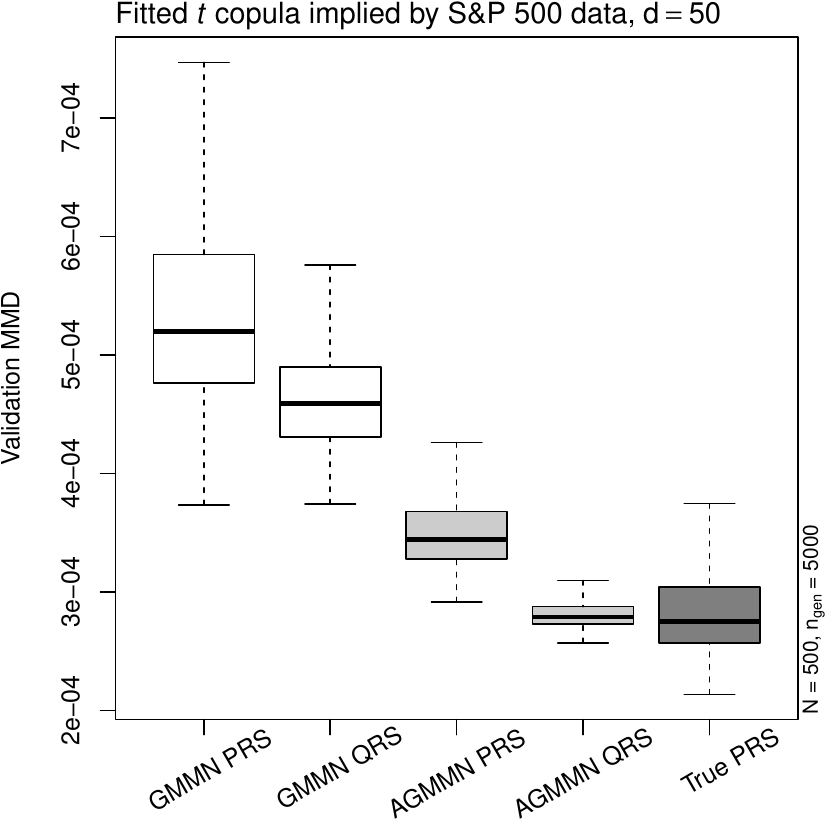}\hspace{12mm}
  \includegraphics[width=0.355\textwidth]{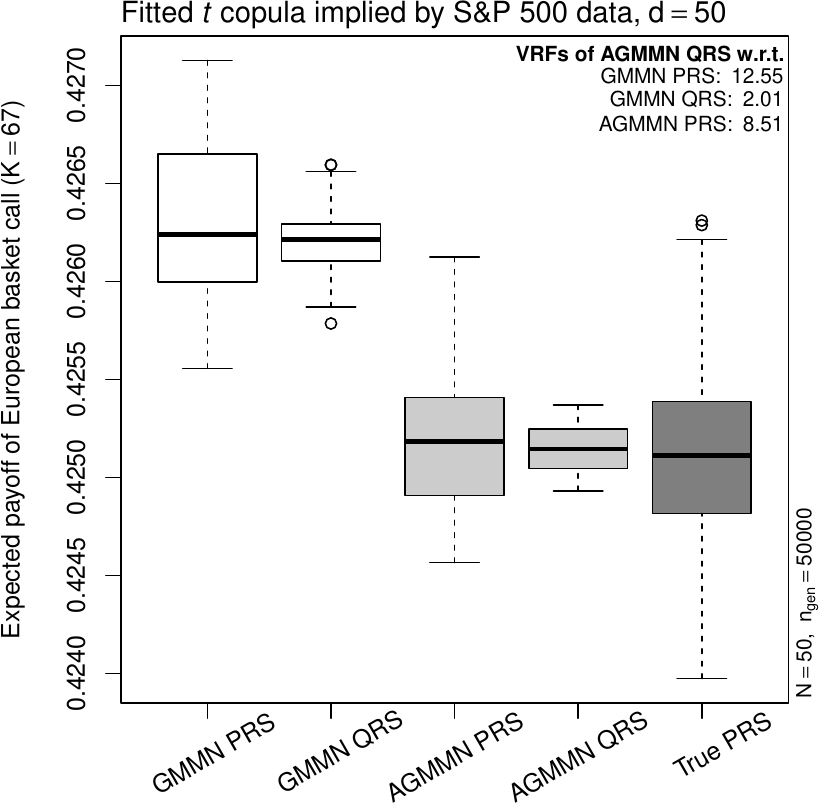}\\[2mm]
  \includegraphics[width=0.8\textwidth]{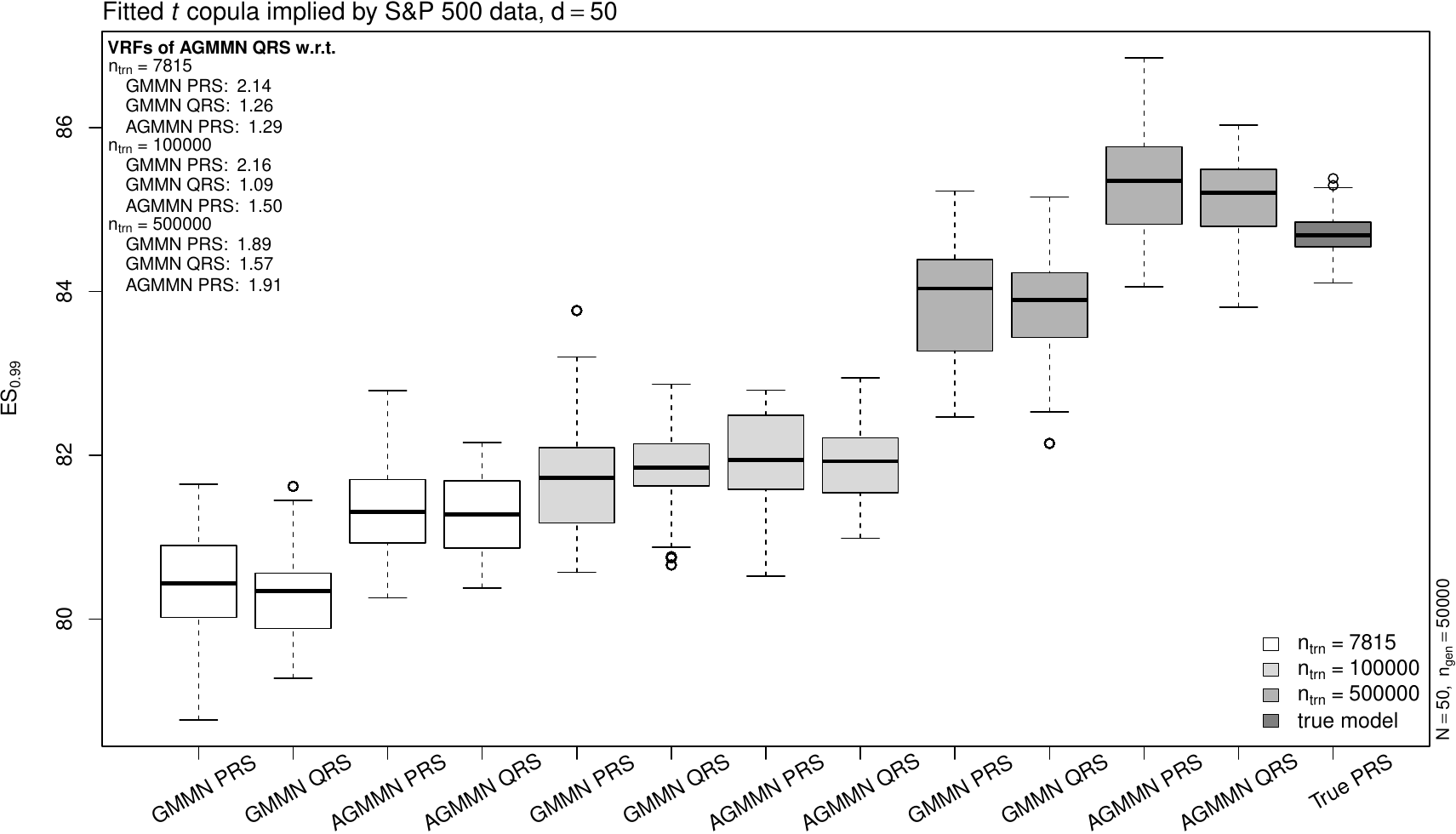}
  \caption{Top left: Boxplots of $N = 500$ realizations of the validation
    MMD~\eqref{eq:MMD} based on $\hval$ in~\eqref{eq:hval} computed between a
    sample from a $t$ copula (fitted to the pseudo-observations of the
    standardized residuals of 50 constituents of the S\&P~500 from 1985-01-01 to
    2015-12-31 after deGARCHing) and a pseudo-random sample (PRS) of a trained
    GMMN, a quasi-random sample (QRS) of a trained GMMN, a PRS of a trained
    AGMMN, a QRS of a trained AGMMN, and a pseudo-random sample from the fitted
    $t$ copula (labeled ``True PRS''), all of size $\ngen = 5000$. Top right:
    Boxplots of $N = 50$ realizations of the GMMN MC estimator $\muGMMNMC$
    (labeled ``GMMN PRS''), the GMMN RQMC estimator $\muGMMNRQMC$ (labeled
    ``GMMN QRS''), the AGMMN MC estimator $\muAGMMNMC$ (labeled ``AGMMN PRS''),
    the AGMMN RQMC estimator $\muAGMMNRQMC$ (labeled ``AGMMN QRS'') and the
    copula MC estimator $\mucopMC$ (labeled ``True PRS'') of the expected payoff
    of a basket call option, using $\ngen = 50\,000$. Bottom: Boxplots of
    $N = 50$ realizations of $\muGMMNMC$, $\muGMMNRQMC$, $\muAGMMNMC$,
    $\muAGMMNRQMC$ and $\mucopMC$ (the latter in dark gray) of the expected shortfall
    $\ES_{0.99}(S)$, where models are fitted with training sample sizes
    $\ntrn = 7815$ (white), $\ntrn = 100\,000$ (light gray), and
    $\ntrn = 500\,000$ (medium gray), using $\ngen = 50\,000$.}
  \label{fig:sim1}
\end{figure}

We first assess the fit of the trained GMMNs and AGMMNs. The top left plot of
Figure~\ref{fig:sim1} displays boxplots of $N = 500$ realizations of the
validation MMD~\eqref{eq:MMD} based on $\hval$ in~\eqref{eq:hval}, computed with
respect to a GMMN pseudo-random sample (labeled ``GMMN PRS''), a GMMN quasi-random
sample (labeled ``GMMN QRS''), an AGMMN pseudo-random sample (labeled ``AGMMN
PRS''), an AGMMN quasi-random sample (labeled ``AGMMN QRS'') and a sample from the
true underlying fitted $t$ copula (labeled ``True PRS'').
We see that in comparison to GMMN-generated samples, AGMMN-generated samples
produce validation MMD values closer to those produced by the true samples.
Furthermore, AGMMN quasi-random samples achieve validation MMD values that
are most concentrated around and most similar to those of the true underlying
copula. From the validation MMD metric's perspective, AGMMNs achieve better fits
than GMMNs, and capture the model's target distribution sufficiently well.

An interesting question is whether the MC and RQMC estimators of the two quantities $\mu$
of interest support the AGMMNs' superiority observed above in terms of the
validation MMD. First, let us focus on the expected payoff of the basket call option.
The top right plot of Figure~\ref{fig:sim1} presents boxplots of
$N = 50$ realizations of the GMMN MC estimator $\muGMMNMC$ (labeled ``GMMN
PRS''), the GMMN RQMC estimator $\muGMMNRQMC$ (labeled ``GMMN RQMC''), the AGMMN
MC estimator $\muAGMMNMC$ (labeled ``AGMMN PRS''), the AGMMN RQMC estimator
$\muAGMMNRQMC$ (labeled ``AGMMN RQMC'') and the true copula MC estimator
$\mucopMC$ (labeled ``True PRS'') of the expected payoff of the basket call
option. We also report the \emph{variance-reduction factor (VRF)} of
$\muAGMMNRQMC$ with respect to the other estimators, which is given by
\begin{align*}
	\VRF(\mudot) = \frac{\hat{s}^2(\mudot)}{\hat{s}^2(\muAGMMNRQMC)},
\end{align*}
where $\hat{s}^2(\mudot)$ stands for the sample variance
of $\mudot \in \{\muGMMNMC, \muGMMNRQMC, \muAGMMNMC\}$. We
observe that both $\muAGMMNMC$ and $\muAGMMNRQMC$ can give accurate estimates while
neither $\muGMMNMC$ nor $\muGMMNRQMC$ can. And $\muAGMMNRQMC$ has the lowest variance
among all estimators considered, so is the most precise estimator.

Now consider expected shortfall. The white boxplots in the bottom plot of
Figure~\ref{fig:sim1} depict the distribution of estimates of the expected
shortfall $\ES_{0.99}(S)$ given by the same set of models as before. This time,
all four estimators fail to give estimates close to that of $\mucopMC$ (dark
gray, on the right). Noting that $\ES_{0.99}(S)$ is much more sensitive to
the shape of the tail of the target distribution than the expected payoff of the
basket call option, this to some extent implies that the training sample size
$\ntrn = 7815$ is too small for GMMNs and AGMMNs to sufficiently learn the tail
of the underlying 50-dimensional distribution. To support this claim, we also
include estimates given by GMMNs and AGMMNs that are trained on samples of sizes
$\ntrn = 100\,000$ (light gray) and $\ntrn = 500\,000$ (medium gray). We observe
that all neural network based estimators approach the true copula MC estimator
as $\ntrn$ increases and that the AGMMN estimators consistently outperform the
GMMN estimators in terms of both bias and variance for each $\ntrn$. For
$\ntrn = 500\,000$, $\muAGMMNMC$ and $\muAGMMNRQMC$ are able to give accurate
estimates while $\muGMMNMC$ and $\muGMMNRQMC$ are still biased. In conclusion,
we see evidence that due to the improved fit, AGMMNs lead to lower bias and
variance than GMMNs also for the quantities $\mu$ of interest we considered.

\section{A real data example: S\&P~500 and FTSE 100 constituents}\label{sec:prediction}
In this section, we assess the performance of AGMMNs in comparison to GMMNs on
two real-world datasets, the first one being the S\&P~500 constituent data
specified in Section~\ref{sec:simulation}, and the other consisting of 6841
daily adjusted closing prices of 50 constituents of the Financial Times Stock
Exchange 100 (FTSE) from 1988-01-01 to 2015-12-31. The latter data are available
in the \R\ package \texttt{qrmdata} (\cite{qrmdata}); see Appendix~A.2 for
details. For both datasets, we follow the same deGARCHing procedure as described
in Section~\ref{sec:simulation} to obtain pseudo-observations of marginal
standardized residuals. We use the first $\ntrn = 6000$ of the 7815 observations
of the S\&P~500 dataset as training data for all models, and the remaining
$\ntst=1815$ as test data to evaluate and compare all models. The FTSE~100
dataset has 6841 observations, and we take the first $\ntrn = 5500$ as training
data and the remaining $\ntst=1341$ as test data. In what follows, we use
$\ndat\in\{\ntrn,\ntst\}$ to denote the sample size of the training or test data,
depending on which part of the datasets we consider. Training of all models for both datasets is conducted in mini-batches with $\nbat = 2000$.

We train GMMNs and AGMMNs with the following choices of architectures. As basic
models, we consider GMMNs and AGMMNs with one hidden layer of size 300, the
architecture used in all experiments so far. Such models are denoted by
``$\text{G}$'' and ``$\text{A}$'', respectively. We also consider GMMNs and
AGMMNs with a single wide hidden layer of size 10\,000, denoted by
``$\text{G}_{\text{w}}$'' and ``$\text{A}_{\text{w}}$'', respectively. And we
consider GMMNs and AGMMNs with a deeper architecture that consists of two hidden
layers of sizes 2000 and 500, abbreviated as ``$\text{G}_{\text{d}}$'' and
``$\text{A}_{\text{d}}$'', respectively. To incorporate the parameter
uncertainty introduced by the stochastic nature of the training procedure of
neural network models, each training is repeated three times and we report the
respective results separately. For comparison, we also include two
parametrically estimated copulas, a $t$ copula (labeled ``$t$'') and an R-vine
copula (labeled ``V''); the former is fitted via the \R\ package \texttt{copula}
with the fitted degrees of freedom being 32.98 for the S\&P~500 dataset and
16.47 for the FTSE 100 dataset, and the latter is fitted via the \R\ package
\texttt{VineCopula} where all bivariate copula families available in the package
are considered for the fitting.

This time around, we evaluate the models' performances with an average
Cram\'er-von-Mises type test statistic (\cite{remillardscaillet2009}) given by
\begin{align*}
  \ACvM = \frac{1}{\nrep}\sum_{k=1}^{\nrep} \Biggl( \frac{1}{\sqrt{\frac{1}{\ndat} + \frac{1}{\ngen}}} \int_{[0,1]^d}\bigl(C_{\ndat}(\bm{u}) - C^{\,\bigcdot}_{\ngen, k}(\bm{u}) \bigr)^2 \rd\bm{u}\Biggr),
\end{align*}
where $C_{\ndat}$ is the empirical copula of $\ndat$ pseudo-observations for one
of the two datasets under consideration and $C_{\ngen, k}^{\,\bigcdot}$ is the
empirical copula of $\ngen$ samples generated from model
$C_{\ngen, k}^{\,\bigcdot}\in\{C_{\ngen, k}^{\text{G}_{\text{w}}},C_{\ngen,
	k}^{\text{A}_{\text{w}}},C_{\ngen, k}^{\text{G}_{\text{d}}},C_{\ngen,
	k}^{\text{A}_{\text{d}}},C_{\ngen, k}^{t},C_{\ngen, k}^{\text{V}}\}$ in
replication $k=1,\dots,\nrep$. In all cases, we use $\nrep = 25$ replications
and choose the sample size $\ngen = \ndat$.  As mentioned before, $\ndat$ stands
for $\ntrn$ or $\ntst$, and we indeed compute the $\ACvM$ statistic for both the
training data (\emph{training data ACvM}) and the test data (\emph{test data
	ACvM}), to assess the prediction quality of all estimated models (via the test
data ACvM) against the goodness-of-fit of all estimated models (via the training
data ACvM). This allows us to investigate whether improved training (x-axis) translates
to improved prediction (y-axis).

\begin{figure}[htbp]
  \includegraphics[width=0.48\textwidth]{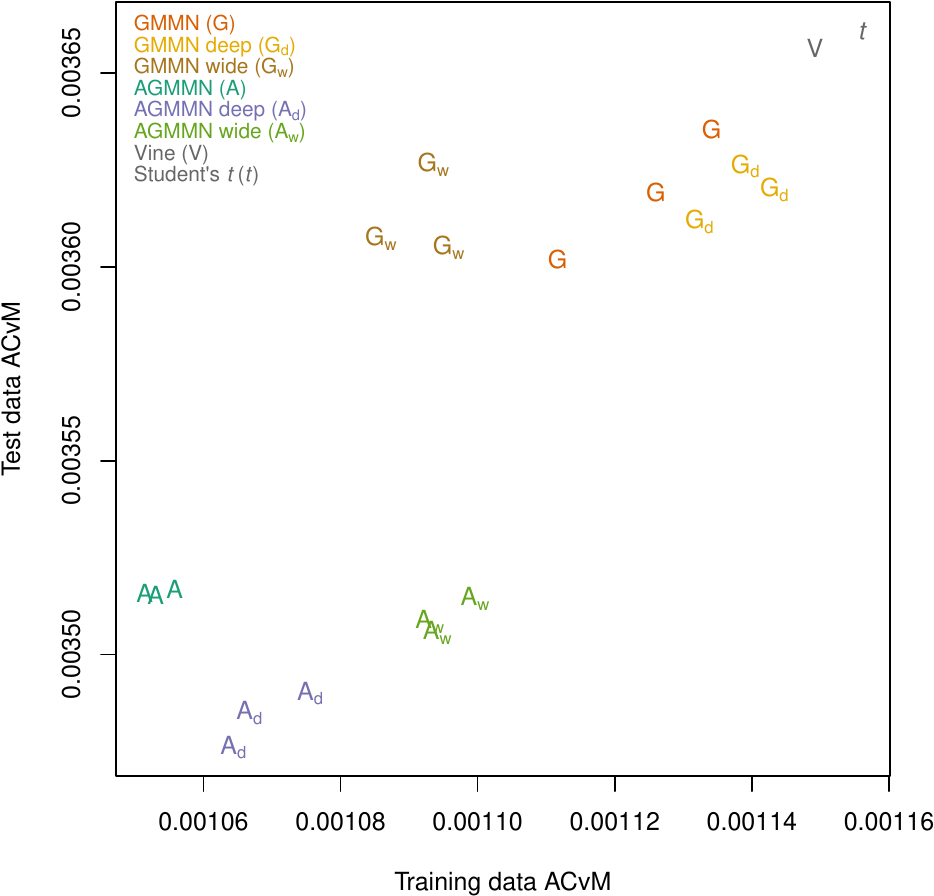}
  \hfill
  \includegraphics[width=0.46\textwidth]{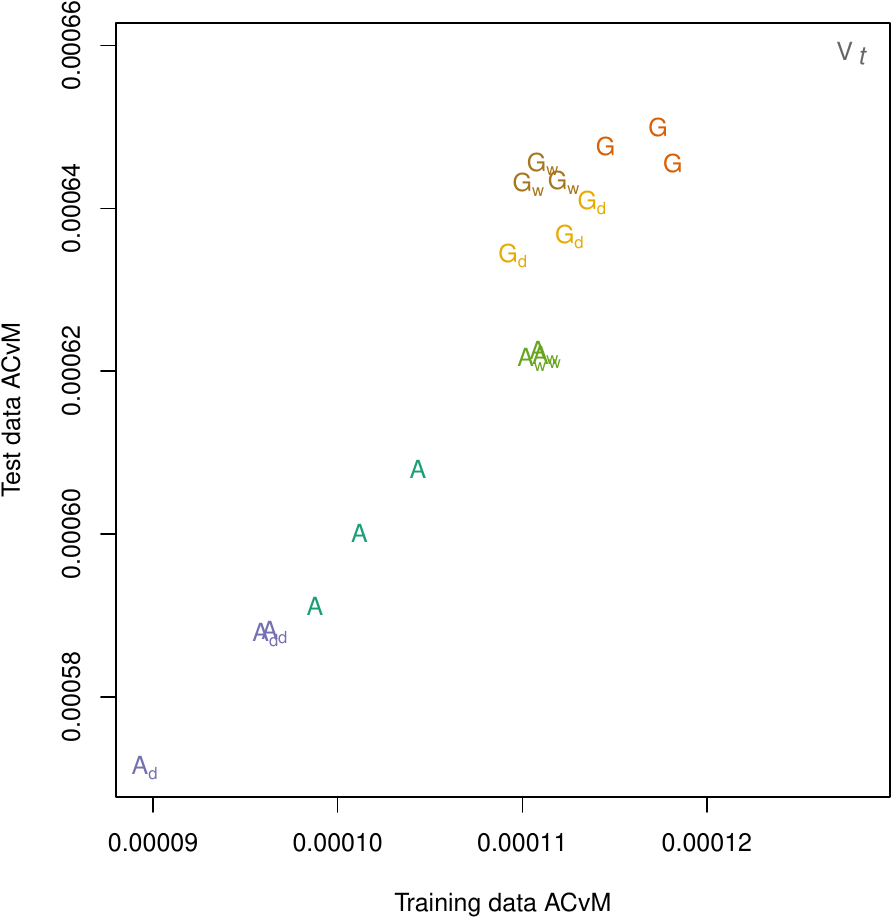}
  \caption{Test data ACvM vs training data ACvM of the S\&P~500 data (left) and
    the FTSE~100 data (right).}\label{fig:prediction_part_nonNNs}
\end{figure}

Figure~\ref{fig:prediction_part_nonNNs} presents plots (left: S\&P~500 constituent
data; right: FTSE~100 constituent data) of the test data ACvM against the training
data ACvM for each model and, in case of neural network models, the three replicates.
Although GMMNs can sometimes achieve a training data ACvM comparable to that of
AGMMNs, we see that AGMMNs reach a better test data ACvM, and this regardless of the
architecture. Furthermore, while the more complex deep and wide architectures
still do not make GMMNs competitive in terms of their prediction quality, AGMMNs can
take advantage of the deep architecture and further improve their prediction
performance. A possible explanation for this result is that the learning of
GMMNs is not efficient enough for exploiting more complex architectures, but the
adaptive learning procedure underlying AGMMNs is.

\section{Conclusion}\label{sec:concl}
We introduced an adaptive bandwidth selection procedure for the mixture kernel
in the maximum mean discrepancy (MMD) to enhance the training of generative
moment matching networks (GMMNs) for improved learning and predictive modeling
of dependence structures. By dynamically increasing the number of kernels during
training, guided by the relative error of the training loss and by selecting the
corresponding bandwidths via the median heuristic, the resulting adaptive GMMNs
(AGMMNs) improved GMMNs with fixed bandwidths notably. AGMMNs achieved a better
performance in terms of validation MMD trajectories, generated samples, and
validation MMD values, while the training procedure is sufficiently simple to
implement and training time is roughly comparable to that of GMMNs. Concerning
the latter, the main increase in run time for training AGMMNs is due to the MMD
being based on a larger number of kernels after updates, as well as the
additional computation of the validation loss once per epoch. However, this
increased computational cost is roughly offset by the early stopping criterion
we introduced based on the relative error of the already computed validation
loss.

More specifically, we demonstrated the superiority of AGMMNs over GMMNs and
classical parametric copulas in three key applications. First, AGMMNs enabled
efficient quasi-random sampling from copulas in dimensions as high as 100 for the
first time, which we showed using as functionals of interest the expected shortfall
(a risk measure from quantitative risk management), the expected shortfall
contribution (a capital allocation principle from quantitative risk management),
and the expected payoff of a European basket call option (an example from finance).
We identified the problem that the underlying randomized Sobol' sequence
deteriorates already in fairly low dimensions and addressed it by choosing a
smaller neural network input layer dimension (10-dimensional prior distribution)
than the output layer dimension (100-dimensional). This comes at the cost of
introducing a bias which, however, is of smaller order than for GMMNs. Second,
using a copula fitted to the standardized residuals of 50 constituents of the
S\&P~500 after deGARCHing, we showed that AGMMNs outperform GMMNs in terms of
replicated validation MMDs, as well as in Monte Carlo and quasi-Monte Carlo
applications based on the expected payoff of a basket call option and expected
shortfall. Finally, on 50 constituents of both the S\&P~500 (same data as before)
and the FTSE~100, we demonstrated that the improved training of AGMMNs translates
to an improved model prediction of AGMMNs over both GMMNs and typical parametric
copulas used in that context.

One possible future research direction is to train with smaller training sample
sizes. Depending on the application of interest, we suspect that some
applications allow one to largely reduce the training sample size while still
more than adequately capturing the dependence, which would be important for
practical applications. Another research direction would be to investigate the
effect of different transformations of neural network generated data on the
quality of the learned distributions. What seems to have been missed by
\cite{jankeghanmisteike2021} is the fact that we (also in
\cite{hofertprasadzhu2021a}, \cite{hofertprasadzhu2022a},
\cite{hofertprasadzhu2023a}, \cite{hofertprasadzhu2023b}) do not simply use the
samples produced by the sigmoid activation function in the output layer as
copula samples (as there would indeed be no guarantee of at least approximately
having standard uniform margins), but the pseudo-observations thereof, which
guarantee $\U(\{\frac{1}{\ngen+1},\dots,\frac{\ngen}{\ngen+1}\})$
margins. Another idea would be draw from beta distributions based on the $\ngen$
generated ranks as done when sampling empirical beta copulas; we did not see the
need to go in this direction in the current work since our $\ngen$ was typically
large, but for smaller $\ngen$, this approach may be preferred.

\subsection*{Supplementary material}
Appendix~A.1 contains a simulation of the fluctuation of points of a
digital-shifted Sobol' sequence as the dimension increases. Appendix~A.2
contains details of the preprocessing of the data, as well as the list of
constituents of the S\&P~500 dataset used in Sections~\ref{sec:simulation} and
\ref{sec:prediction}, and the FTSE~100 dataset used in Section~\ref{sec:prediction}.

\subsection*{Acknowledgements} %
The computations were performed using the research computing facilities provided by
Information Technology Services (ITS) at The University of Hong Kong. The
authors gratefully acknowledge ITS for its technical support and assistance
throughout this work.

\subsection*{Data availability statement}
The data that support the findings of this study are available in the public domain
resources cited.

\subsection*{Disclosure statement} %
The authors have no potential competing interest to declare.

\printbibliography[heading=bibintoc]

\end{document}

